\pdfoutput=1

\documentclass[11pt]{article}

\usepackage{acl}
\usepackage{amsmath}
\usepackage{bm}
\usepackage{amsfonts}
\usepackage{booktabs}
\usepackage{multirow}
\usepackage{times}
\usepackage{latexsym}
\usepackage{cleveref}
\crefformat{section}{\S#2#1#3} 
\crefformat{subsection}{\S#2#1#3}
\crefformat{subsubsection}{\S#2#1#3}
\date{\vspace{-5ex}} 

\usepackage[T1]{fontenc}

\usepackage[utf8]{inputenc}
\usepackage{times}
\usepackage{pifont}
\usepackage{latexsym}
\usepackage{graphicx} 
\usepackage{natbib}  
\usepackage{caption} 
\usepackage{amssymb}
\usepackage{color}
\usepackage{subfigure}
\usepackage{amsmath}
\usepackage{amsthm}
\usepackage{multicol}
\usepackage{multirow}
\usepackage{booktabs}
\usepackage{microtype}
\usepackage{macros}
\usepackage{comment}
\usepackage{soul}
\usepackage{framed}
\usepackage{titlesec}
\usepackage[inline]{enumitem}
\usepackage{hyperref}

\definecolor{blueish}{RGB}{31, 78, 192}
\definecolor{orangeish}{RGB}{240, 147, 41}
\definecolor{mypink1}{rgb}{0.858, 0.188, 0.478}
\definecolor{mypink2}{RGB}{219, 48, 122}
\definecolor{mypink3}{cmyk}{0, 0.7808, 0.4429, 0.1412}
\definecolor{mygray}{gray}{0.6}
\definecolor{darkbluee}{RGB}{0,17, 113}

\newtheorem{proposition}{Proposition}

\newcommand{\lmflatness}{\textsc{pFlat}}

\newcommand{\sensitivity}{\textsc{Sen}}
\newcommand{\mi}{\textsc{MI}}
\newcommand{\rate}{{\tt Rate}}

\newcommand{\subscript}[2]{\textcolor{darkbluee}{$#1 _ #2$}}

\title{
\vspace*{-0.5in}
{{\small \hfill EMNLP-Findings'23}\\
\vspace*{.25in}} 
Flatness-Aware Prompt Selection \\ Improves Accuracy and Sample Efficiency}

\author{Lingfeng Shen$^\heartsuit$\, \hfill Weiting Tan$^\heartsuit$ \hfill Boyuan Zheng \hfill Daniel Khashabi\\
  Center for Language and Speech Processing \and Computer Science Department\\Johns Hopkins University, Baltimore MD \\
  \texttt{\{lshen30, wtan12, bzheng12, danielk\}@jhu.edu}
}

\newcommand{\dataset}{\mathcal{D}}
\newcommand{\datasetX}{\mathcal{D}_{X}}

\newcommand{\x}{\boldsymbol{x}}
\newcommand{\y}{\boldsymbol{y}}
\newcommand{\prompts}{\mathcal{P}}
\begin{document}
\maketitle
\begin{abstract}

\blfootnote{$\heartsuit$ Equal contribution.}

With the growing capabilities of large language models, prompting them has become the dominant way to access them. 
This has motivated the development of strategies for automatically selecting effective language prompts. 
In this paper, we introduce \lmflatness~ (prompt flatness), a new metric to quantify the expected utility of a language prompt. This metric is 
inspired by \emph{flatness} regularization in statistical learning that quantifies the robustness of the model towards its parameter perturbations.  
We provide theoretical foundations for this metric and its relationship with other prompt selection metrics, providing a comprehensive understanding of existing methods. 
Empirically, we show that
combining \lmflatness{} with existing metrics
improves
{
both performance and sample efficiency. 
Our metric outperforms the previous prompt selection metrics with an average increase of 10\% in Pearson correlation across 6 classification benchmarks, and the prompt selected by our metric gains 5\% higher accuracy than previous metrics across the benchmarks.}\footnote{
The code is accessible {here}: \url{https://github.com/shadowkiller33/flatness}.
}
\end{abstract}

\section{Introduction}\label{intro}

Manually ``engineering'' prompts for large language models (LLMs) have been shown to lead to tremendous performance gains and have been a subject of intense study in recent years  
\cite{schick2021exploiting,reynolds2021prompt,mishra2022reframing}. However, the task of prompt engineering can be challenging due to the difficulty in determining the effectiveness of a prompt solely based on its raw text form. Consequently, this process is typically carried out manually, which can be laborious and time-intensive. In particular, LLMs may produce vastly different predictive distributions for two seemingly comparable prompts, despite their semantic similarity~\cite{mishra2022reframing}. This phenomenon results in an unexpectedly high level of variability.~\cite{DBLP:journals/tacl/JiangXAN20,perez2021true,DBLP:journals/tacl/ElazarKRRHSG21}.

\begin{figure}[t]
    \centering
    \includegraphics[scale=1.18,trim=0.2cm 0.01cm 0cm 0.1cm]{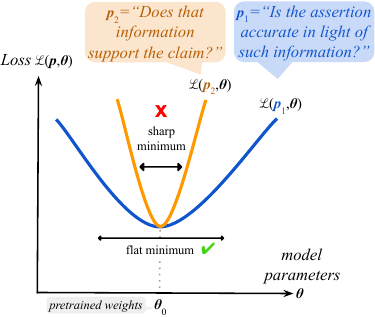}
    \caption{
        We show that \emph{prompt flatness} is an effective indicator of a prompt's performance on an LLM.
        For example, if two prompts \textcolor{blueish}{$p_{1}$}, \textcolor{orangeish}{$p_{2}$} incurs the same loss on an LLM parameterized by $\theta_0$, i.e., $\mathcal{L}(\textcolor{blueish}{p_{1}}, \theta_0)=\mathcal{L}(\textcolor{orangeish}{p_{2}}, \theta_0)$, we find that the one with a flatter loss landscape of LLM parameters (\textcolor{blueish}{$p_{1}$}, in this visualization) is better. 
    }
    \label{fig1}
\end{figure}

In response to such difficulties,
recent works propose metrics for automatic prompt selection. 
Notably, \citet{sorensen2022information} introduces \emph{Mutual Information (MI)} to quantify the shared information between prediction and inputs. Further, \citet{chen2022relation} introduces \emph{Sensitivity (\sensitivity)} to quantify model receptiveness to textual perturbations of the input prompts. 
Despite such metrics' empirical effectiveness, the underlying principles that enable them are not well understood.

This motivates the following questions:  
\begin{enumerate*}[label=(\subscript{RQ}{{\arabic*}}),leftmargin=35pt]
    \itemsep-0.6em 
    \item \label{Q1}  
    What makes the existing methods for prompt selection effective? 
    \item \label{Q2} How are these existing methods connected?
    \item \label{Q3} 
    Are there any new metrics complementary to the existing ones?
\end{enumerate*}

To address the questions above, we study existing methods
from an optimization perspective.
The objective $\mathcal{L}(p, \dataset, \theta_0)$  quantifies the performance of an LLM (parameterized by $\theta_0$) 
on
labeled data $\data$ and a prompt $p$ 
appended to the dataset inputs.
Prompt selection is in effect an optimization on $\mathcal{L}(p, \dataset, \theta_0)$ as a function of different choices of $p$. 
The challenge is that, in practice, there are few labeled data $\dataset$~\cite{perez2021true}, which would make $\mathcal{L}(.)$ an unreliable measure for selecting effective prompts. 
We show that the existing prompt selection metrics (MI and \sensitivity) \cite{sorensen2022information,chen2022relation} approximate the objective function $\mathcal{L}$, and therefore, 
act as its \emph{surrogates}. 
This addresses \ref{Q1} and \ref{Q2} above. 

Additionally, 
to address \ref{Q3} we borrow ideas from statistical learning on flatness-aware optimization~\cite{hochreiter1994simplifying,keskar2017sharpminima}. We introduce \emph{Prompt Flatness} (\lmflatness), a metric that quantifies $\mathcal{L}$'s sensitivity to small perturbations in LLMs parameters, when conditioned on a prompt (see \autoref{fig1} for intuitions).
Our results indicate that prompts with higher flatness generally lead to better accuracy. 

Our formal derivations also show that \lmflatness{} is distinct from and complementary to prior metrics such as MI and \sensitivity. 
Our empirical results (\S\ref{sec:experiments}) on six classification benchmarks and four different model sizes also confirm our theoretical intuition. 
For example, combining \lmflatness\ and MI improves the downstream performance by 6\% accuracy over the prompts selected by MI only. 
Similarly, combining \lmflatness\ and \sensitivity{} improves the downstream performance by 9\% accuracy over prompt selected by \sensitivity{} only. 
Additionally, using \lmflatness\ substantially improves sample efficiency, an important feature of low-resource scenarios. 

In summary, our contributions are: 
(a) We propose a formal optimization framework that unifies several existing prompt selection metrics such as MI and \sensitivity. 
(b) Enabled by our formalism, we introduce \lmflatness{}, a metric for selecting prompts that is more robust to LLMs' parametric perturbations. 
(c) We conduct comprehensive experiments and the results demonstrate the effectiveness of our method for prompt selection.

\section{Prompt Selection via Flatness}
We start by introducing the necessary background and the notational convention (\S\ref{subsec:background}), 
then introduce our 
{proposed metric, \lmflatness}
(\S\ref{subsec:flatness}), followed by a discussion of its relation to other existing prompt selection metrics (\S\ref{subsec:other}). 

\subsection{Background and Setup}
\label{subsec:background}
\paragraph{Notation.} We cast prompt selection into an optimization problem.
We are provided with a pre-trained language model $f$ with parameters $\theta \in \reals^m$ which maps each input natural language instance $\x$ to 
$f_{\theta}(\boldsymbol{x}) \in [0, 1]^{|V|}$, a distribution over the label set $V$.
We are also given input-output pairs $\dataset = \set{(\x, \y)}$, where $\boldsymbol{y}$ is a one-hot label.

\paragraph{Prompt selection.}
Given a language model $f$, we seek to minimize the following empirical risk, also called prompt loss in this paper:
$$
\mathcal{L}(p, \dataset, \theta) = \frac{1}{|\dataset|} \sum_{(x,y) \in \dataset}\ell(f_{\theta}(p\circ \x ),\y), 
$$
where $p\circ x$ is the string combination of a prompt $p$ to input $x$, and $\ell$ is an appropriate loss such as cross-entropy that quantifies the gap between gold label $\y$ and predicted distribution $f_{\theta}(p\circ \x )$.

In the classic machine learning literature, it is customary to minimize empirical risk 
$\mathcal{L}(p, \dataset, \theta)$ with respect to the parameters of the underlying model $\theta$. 
However, 
the recent developments in LLMs
~\cite{radford2019language,brown2020language} have 
{resulted in}
an alternative that involves optimization concerning the choice of prompt $p$: 
\begin{equation}
\hat{p} = \argminof{p \in \prompts} 
    {\mathcal{L}(p, \dataset, \theta)}, 
\label{eq:prompt:selection}
\end{equation}
given a collection of natural language prompts $\prompts$ that are ``engineered'' by domain experts~\cite{Schick2020FewShotTG,schick2021exploiting,mishra2022reframing}


\subsection{Prompt Selection via Flatness}
\label{subsec:flatness}
{
Our work draws inspiration from classic machine learning, where studies have demonstrated that using loss \emph{flatness} in model selection leads to improved performance and generalization
}
~\cite{foret2020sharpness,baldassi2020shaping,zheng2021regularizing,stutz2021relating,andriushchenko2022towards}. 
{In this prior literature, }
the optimization is performed 
{with respect}
to model parameters $\theta$.
{Conversely, in the modern NLP literature, the parameters of LLMs are set once they are pre-trained, and further optimization is achieved through input prompts
}
{As a result, it remains to be seen whether the findings from classic machine learning literature will translate to prompting LLMs.}

\paragraph{Robust prompt selection objective.}
We start 
{with the formal definition of} flatness.
Specifically, the goal is to select parameters that are robust to parameter perturbations: 
\begin{align}
\label{eq:flatness:obj}
    \bar{\mathcal{L}}(p, \dataset, \theta) &= \maxof{\norm{\epsilon} < r}{\mathcal{L}(p, \dataset, \theta + \epsilon)}, \\ 
    \hat{p} &= \argminof{p \in \prompts} \bar{\mathcal{L}}(p, \dataset, \theta), 
\end{align}
where $\epsilon$ is a small perturbation added to model parameters $\theta$.
The inner optimization quantifies the worst-case loss upon a small perturbation of the model parameter from its default value, 
{where the perturbations are contained within 
a small algebraic ball, $\norm{\epsilon} < {r}$}.
The overall objective is a minimax optimization~\cite{zheng2021regularizing,stutz2021relating,baldassi2020shaping} i.e., selecting the best prompt $p$ with the smallest worst loss under small perturbations. 
Note that this is a strict generalization of the standard prompt selection objective in \autoref{eq:prompt:selection}.

\paragraph{Flatness definition.}
Since \autoref{eq:flatness:obj} is a non-trivial saddle-point optimization problem, previous work \cite{penalizinggradientnorm2022zhao,zhang2023gradient} 
has 
{approximated it}
with the gradient norm of loss function:  
\begin{gather}
\label{flat:opt}
\bar{\mathcal{L}}(p, \dataset, \theta)
\approx 
\mathcal{L}(p, \dataset, \theta)
 + \alpha  
\mathcal{F}(p, \dataset, \theta) 
\\
\label{flat}
\mathcal{F}(p, \dataset, \theta) \triangleq \norm{
\nabla_{\theta} 
\mathcal{L}(p, \dataset, \theta)
}_2,
\end{gather}
where $\mathcal{F}(p, \dataset, \theta)$ is the accurate analytical definition of \emph{flatness} the loss function $\mathcal{L}(.)$. 
{Intuitively, it quantifies}
how resilient it is against small perturbations in parameter space $\theta$. 

The calculation of $\mathcal{F}$ requires (1) gradient computation of the loss $\mathcal{L}$ and (2) ground-truth labels which may not be available. 
To circumvent these challenges, we introduce an approximation of $\mathcal{F}$.

\paragraph{An efficient surrogate for flatness.}
Here provide an approximate definition of flatness ($\mathcal{F}$ in \autoref{flat}) that does not depend on instance labels. 
Our new metric, $\lmflatness{}$ quantifies the 
{amount}
of changes in LLM confidence values
{upon}
perturbations in its parameters:  
\begin{align}
\begin{split}
\label{ww}
 & \lmflatness{}(p, \datasetX, \theta) = \\  
 &  \hspace{1.4cm} \frac{1}{|\datasetX|}  \sum _{\x \in \datasetX}  \mathbf{E}_{\epsilon_1, \epsilon_2}  \Big[ g(\epsilon_{1}) - g(\epsilon_{2}) \Big], 
\end{split}
\end{align}
where $g(\epsilon) \triangleq \ell \Big(  f_{\theta}(p\circ \x), f_{\theta + \epsilon}(p\circ \x) \Big)$ and $\epsilon_{1}, \epsilon_{2}$ are sampled from a Gaussian distribution $\mathcal{N}(0,\sigma^2)$ with its variance $\sigma^2$ determining the perturbation magnitude.
Furthermore, $\datasetX = \set{\x}$ refers to the input instances only (no labels).
Intuitively, higher $\lmflatness{}$ means higher sensitivity 
towards 
{perturbation in}
model parameter, indicating that 
{the given input prompt, instances, and the model parameters have formed}
a sharper minimum. 
{The formal connection}
between \lmflatness{} and $\mathcal{F}$ is deferred to \autoref{app:prompt-flatness}.


{
Although the precise computation of $\lmflatness{}$ demands numerous Gaussian samples, practically, approximating it with few samples suffice for a reasonable \lmflatness estimate.
We'll demonstrate this in the experiments (\cref{sec:analysis}).
}

\paragraph{Putting it together.}
Incorporating our \lmflatness{} metric (\autoref{ww}) in robust prompt selection objective (\autoref{flat:opt}) we get the following: 
\begin{equation}\label{all}
    \bar{\mathcal{L}}(p, \dataset, \theta) \approx \mathcal{L}(p, \dataset, \theta)
 + \alpha \cdot \lmflatness{}(p,\dataset, \theta),
\end{equation}
where $\alpha$ is a 
{scalar hyperparameter.}
In our experiments, we select the prompt with the smallest $\bar{\mathcal{L}}$ and show that such prompts have better quality than those selected only by MI or \sensitivity. For emphasis, this equation shows that for robust prompt selection according to $\bar{\mathcal{L}}$, it is not enough to use \lmflatness{} alone. 
It should be used in conjunction to 
$\mathcal{L}$ or its approximations (discussed in the next section). 
{
We show this point empirically in Section~\ref{sec:experiments}.
}{The only reason that our metric is not fully zero-shot is that 
the hyper-parameter $\alpha$ has to be selected according to a few examples of a held-out set.}
\subsection{Relation to Prior Prompt Metrics}
\label{subsec:other}

We show that prompt selection through existing methods such as MI~\cite{sorensen2022information} and \sensitivity~\cite{chen2022relation} 
is approximately equivalent to minimizing prompt loss $\mathcal{L}(p, \dataset, \theta)$, as shown in \autoref{flat}. 
{
Therefore, they can be viewed as surrogates to $\mathcal{L}(.)$. 
} 
{Formally, we provide}
the gap between prompt loss and its surrogates (e.g., MI and \sensitivity)
{which}
is determined by the difference (e.g., KL divergence) between a model's predictions and the ground-truth labels.

{
\paragraph{Mutual Information.}
\citet{sorensen2022information} propose to pick prompts that maximize the mutual information between model input and the prediction. 
}

\begin{proposition}
\label{p1}
Mutual information $\operatorname{MI}(p,\dataset, \theta)$ is a surrogate loss for prompt loss $\mathcal{L}(p,\dataset, \theta)$ with a gap quantitatively defined as follows:
\begin{align*}
    \label{A}
        \operatorname{MI}(p,\dataset, \theta)&- \mathcal{L}(p,\dataset, \theta) \\ = c  + &\frac{1}{|\dataset|}\sum _{(\x, \y) \in \dataset} \operatorname{KL}(f_{\theta}(\x\circ p)||\y),
\end{align*}
where $c$ is a constant $c = H(f_{\theta}(\x\circ p))$ that does not depend on {prompt} $p$. $\operatorname{KL}$ refers to KL divergence.
\end{proposition}

{
\paragraph{Sensitivity.}
Give a prompt $p$, \citet{chen2022relation} utilizes the sensitivity of model prediction towards the textual perturbation in $p$. 
}

\begin{proposition}
\label{p2}
Sensitivity $\sensitivity(p,\dataset, \theta)$ is a surrogate loss for prompt loss $\mathcal{L}(p,\dataset, \theta)$ with a gap defined as follows:
\begin{align*}
\sensitivity(p,\dataset, \theta)&-\mathcal{L}(p,\dataset, \theta) \\= &\frac{1}{|\dataset|}\sum _{(x, y) \in \dataset} \mathbf{E}_{p^{\prime}}\left[ \ell_{01}(f_{\theta}(x \circ p^{\prime}),y)\right],\nonumber
\end{align*}    
where $p^{\prime}$ and $\ell_{01}$ refer to the perturbed prompt and 0-1 loss, and $\mathbf{E}_{p^{\prime}}$ is an expectation (average) over different choices of perturbed prompts $p^{\prime}$
\end{proposition}

The detailed analyses are deferred to \autoref{app:prompt-loss}. These derivations show that selecting prompts based on MI and Sen is approximately selecting the prompts with the smallest prompt loss, which shows their connections and explains why they are effective for prompt selection.


\paragraph{Complementarity to \lmflatness{}.}
A corollary of Proposition~\ref{1},\ref{p2} is that prompt-selection metrics such as MI~\cite{sorensen2022information} and \sensitivity~\cite{chen2022relation} are surrogates for prompt loss, which are complementary to \lmflatness{}, for the purpose of robust prompt selection (\autoref{eq:flatness:obj}). 
To see this, it is enough to go back to \autoref{all}, which shows how robust prompt selection decomposes into \lmflatness{} and $\mathcal{L}$. 
Finally, as we see, $\mathcal{L}$ is approximated by \sensitivity{} and MI, which concludes the argument.

\begin{figure*}[!htbp]
\centering
    \begin{minipage}[b]{.49\linewidth}
        \centering
        \includegraphics[scale=0.28, trim=0cm 1.5cm 0cm 2.8cm]{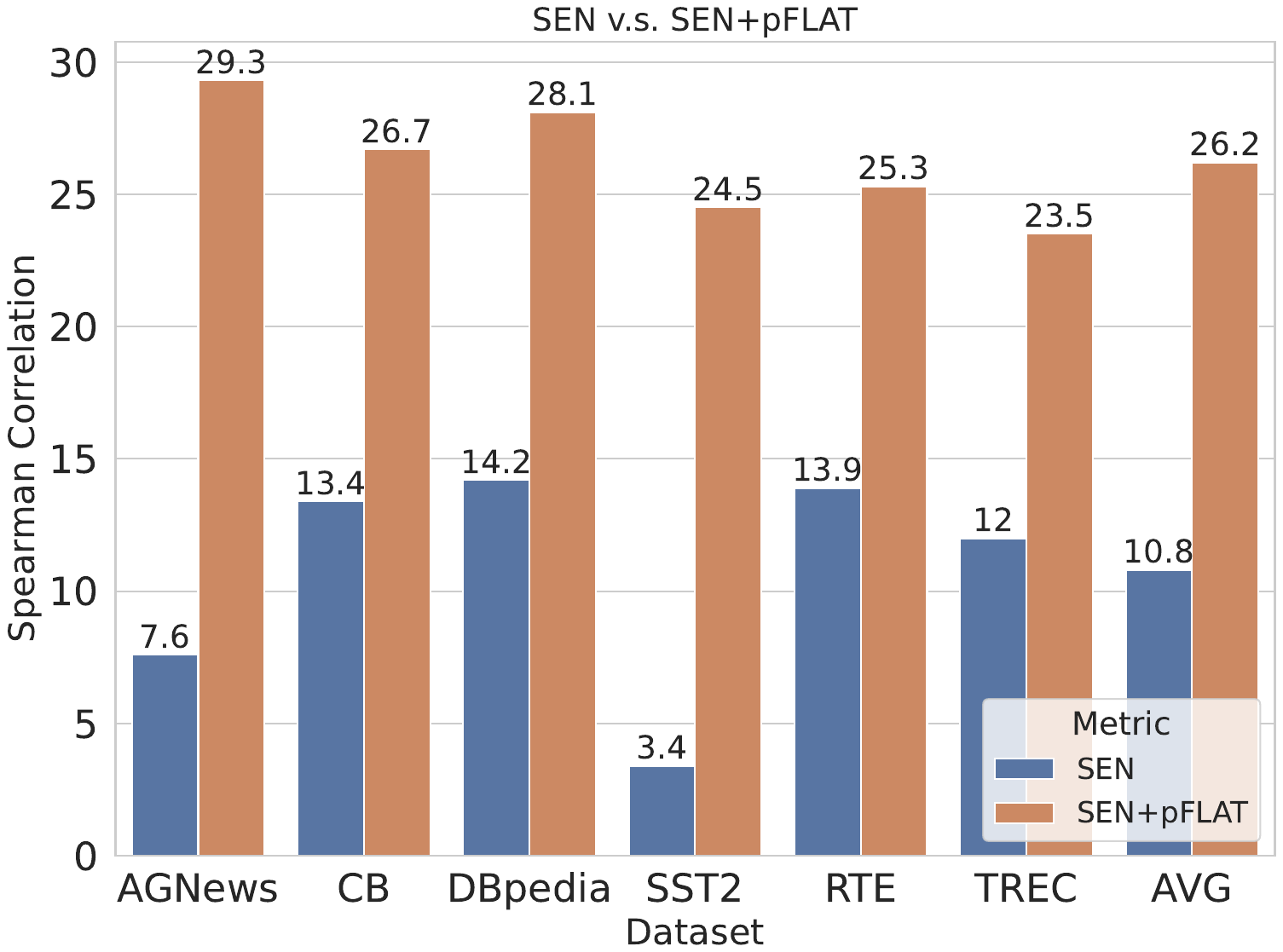}
    \end{minipage}
    \begin{minipage}[b]{.49\linewidth}
        \centering
        \includegraphics[scale=0.28, trim=0cm 1.5cm 0cm 2.8cm]{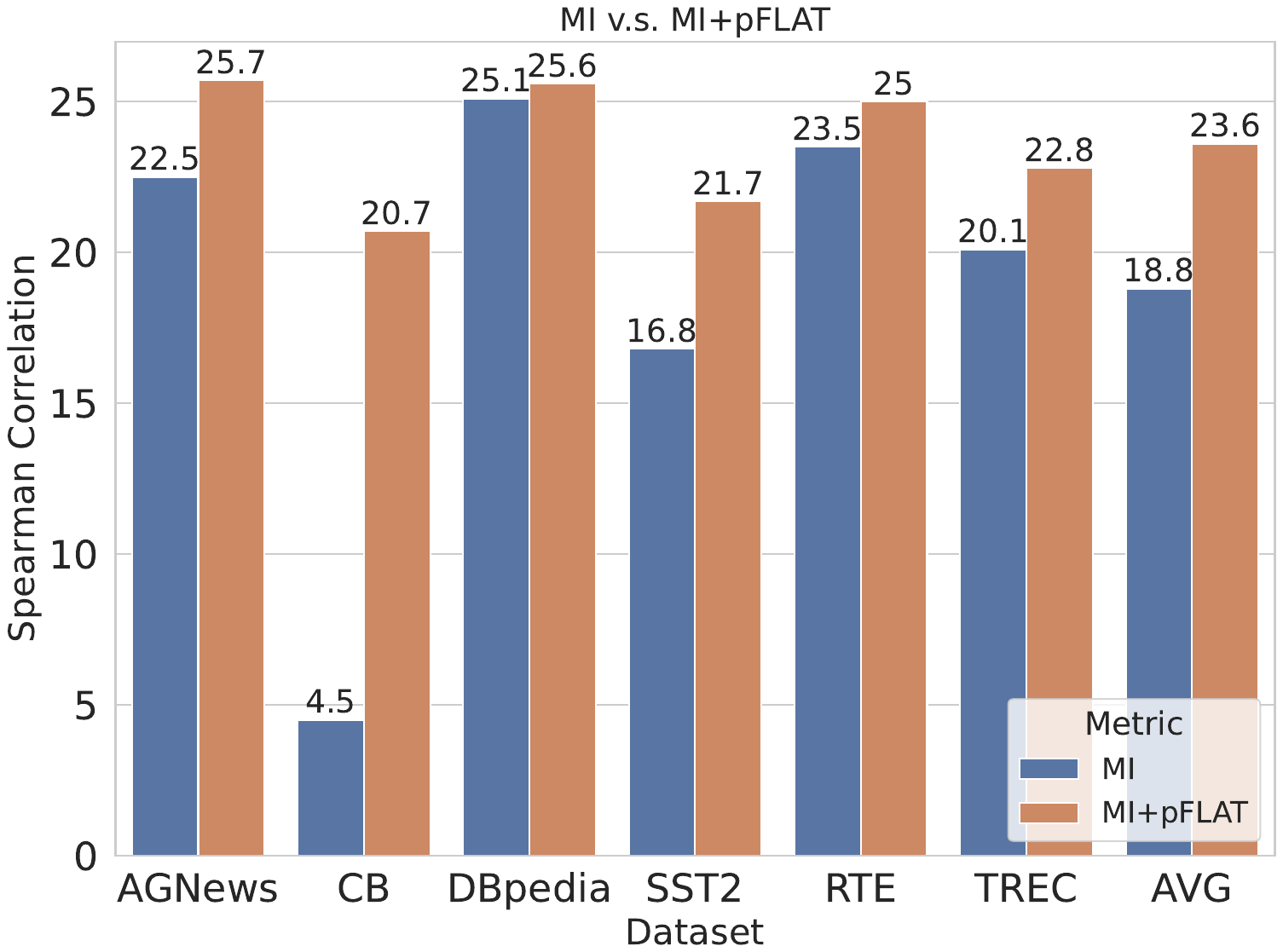}
    \end{minipage}
    \\
    \begin{minipage}[b]{.49\linewidth}
        \centering
        \includegraphics[scale=0.28, trim=0cm 1.5cm 0cm 0.01cm]{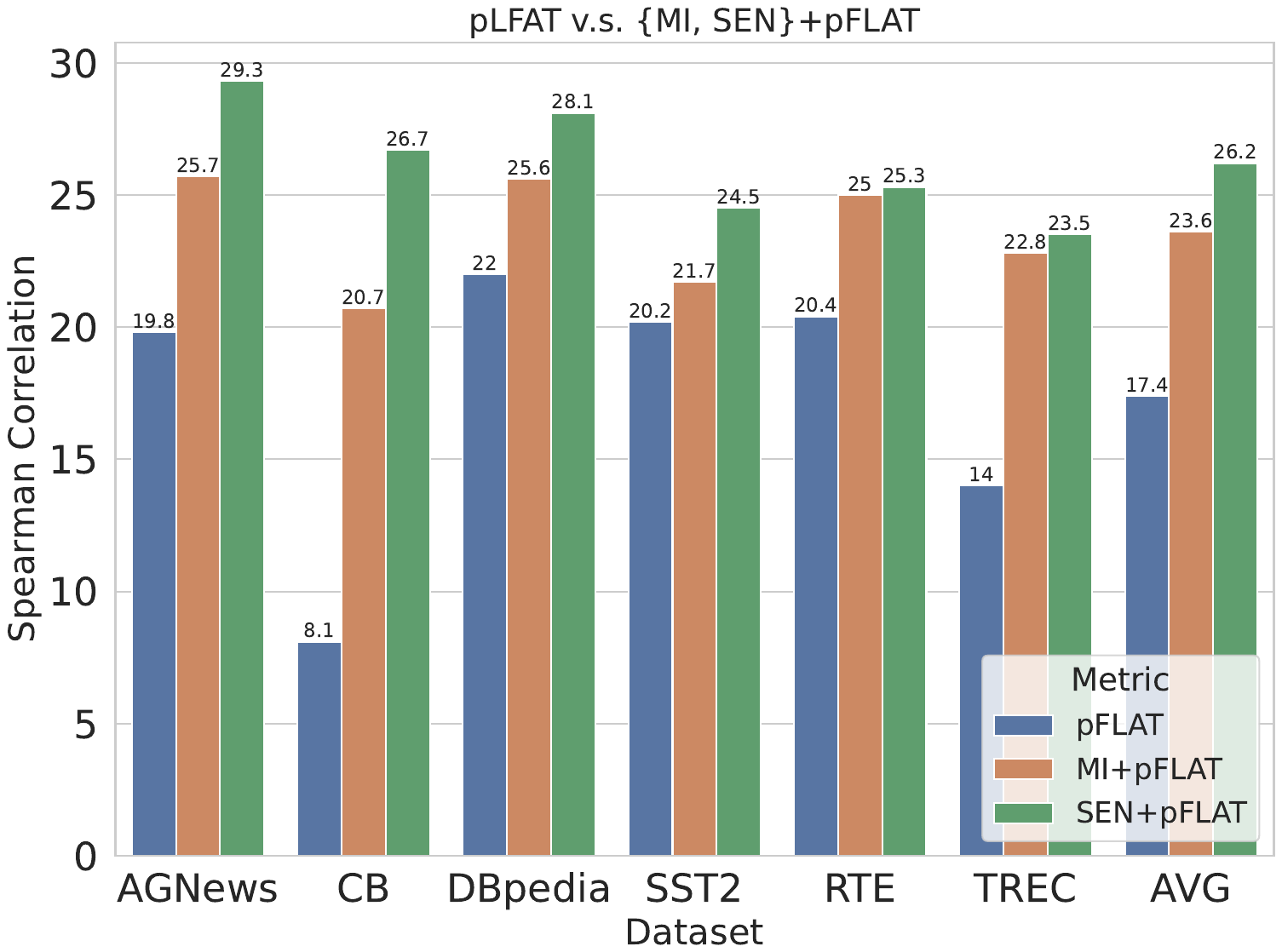}
    \end{minipage}
    \begin{minipage}[b]{.49\linewidth}
        \centering
        \includegraphics[scale=0.28, trim=0cm 1.5cm 0cm 0.01cm]{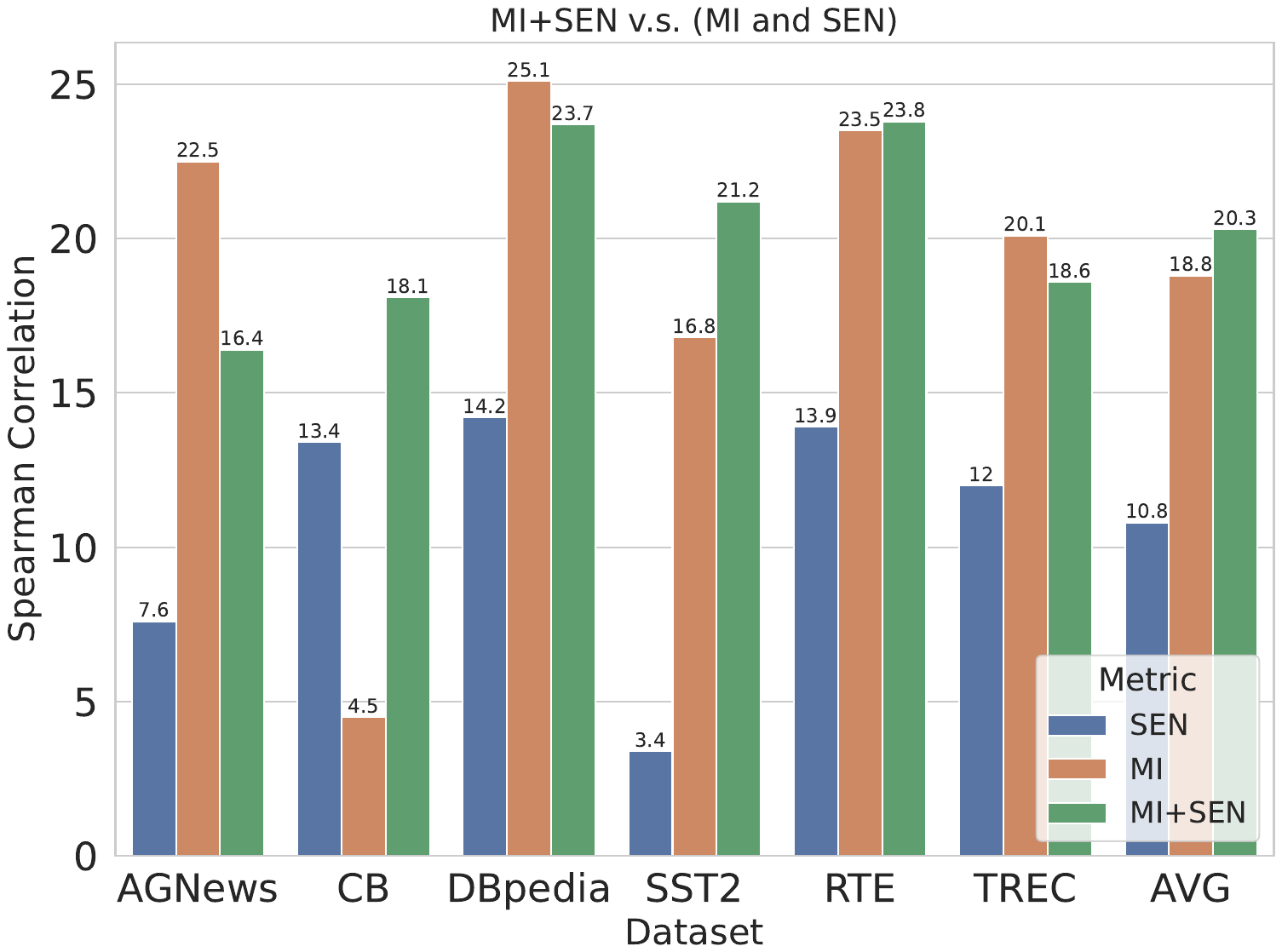}
    \end{minipage}
\caption{Results of correlation evaluation across six datasets and their average (AVG). 
First row: \sensitivity{} vs \sensitivity+\lmflatness{} and \mi{} vs \mi{}+\lmflatness{} show that \textbf{flatness brings consistent improvements over existing metrics}. 
Bottom left: From \lmflatness{} vs \mi+\lmflatness{}, flatness does not perform well when applied alone, as expected. 
Bottom right: \mi+\sensitivity{} vs \mi comparison shows that combining \sensitivity{} and \mi{} brings limited improvement.}
\label{correlation1}
\end{figure*}


\begin{figure*}[ht]
    \centering
    \begin{tabular}{cc}
    \includegraphics[scale=0.293, trim=1.35cm 1.5cm 0cm 0cm]{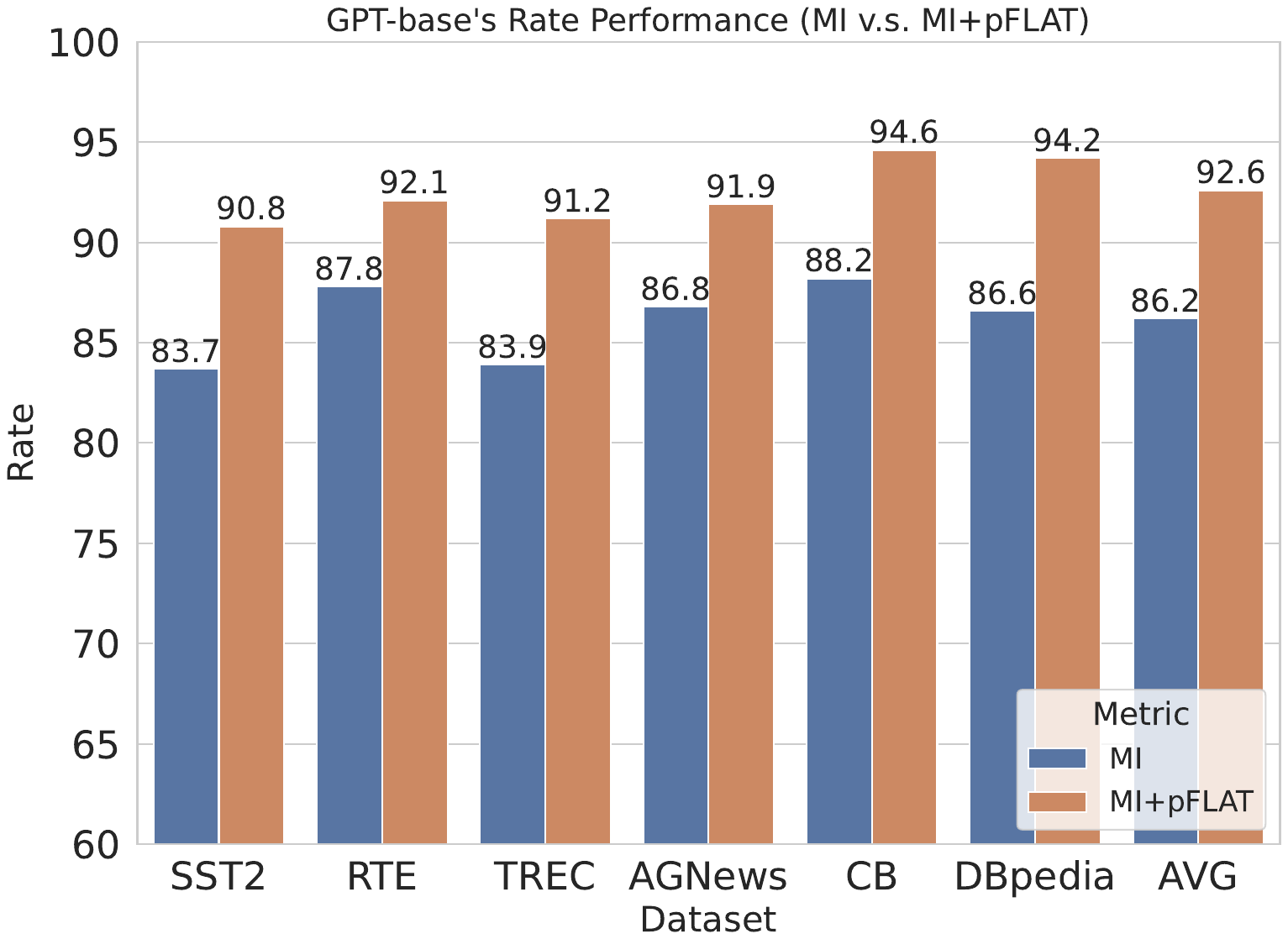}
    \includegraphics[scale=0.293, trim=1.65cm 1.5cm 0cm 0cm]{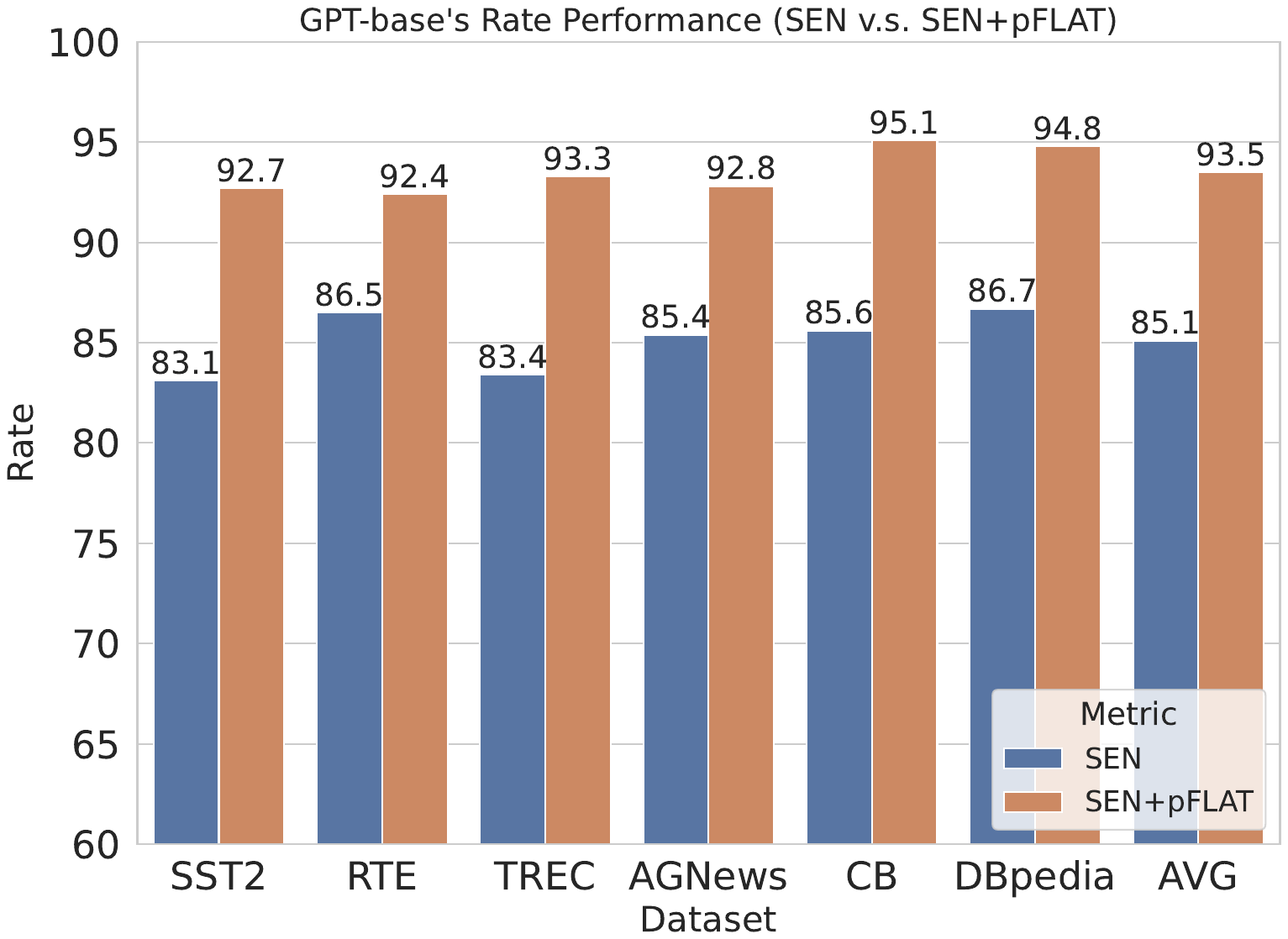}
    \end{tabular}
    \caption{
        \rate (reflecting the ability to select better prompts) computed for prompt selection across six datasets and their average performance, using GPT2-{\tt base} model. 
        We can see that combining \textbf{flatness with existing metrics  (\mi+\lmflatness{} or \sensitivity+\lmflatness{}) is consistently better than not using \lmflatness}. 
        }
    \label{fig2}
\end{figure*}

\section{Experiments}
\label{sec:experiments}


We conduct extensive experiments to assess the effectiveness of prompt selection metrics. 

\paragraph{Experimental setup.}
We experiment with a variety of classification benchmarks: 
AGNews~\cite{zhang2015character}, CB~\cite{de2019commitmentbank}, DBpedia~\cite{zhang2015character}, 
SST-2~\cite{socher2013recursive}, 
RTE~\cite{dagan2005pascal}, 
and TREC~\cite{voorhees2000building}. 
We choose four different GPT-2: {\tt base}, {\tt medium}, {\tt large}, and {\tt xl} sizes.\footnote{
The models are accessible at \url{https://huggingface.co/gpt2}.
}

\paragraph{Held-out set for $\alpha$ hyperparameter.}
For each dataset, we create a small dev-set by randomly selecting 8 labeled sentences per class to tune the value of $\alpha$.

\paragraph{Implementation.}
We prepare 20 human-written instructions by the authors (included in Appendix~\ref{app:discrete-prompt}) appended by random demonstrations for each task.  
The number of demonstrations is set as 5, which matches the settings in \citet{sorensen2022information} for a fair comparison. 
We use $\operatorname{MI}$, $\sensitivity$, $\lmflatness{}$, and their combinations for comparison. 
The results are averaged on three random seeds. {
We estimate $\lmflatness{}$ (\autoref{ww}) via 5 random Gaussian perturbations of LLM parameters with variance  ${\sigma^2}$ set to {\tt 1e-4}. 
Later, we offer an assessment of the influence of this estimation (\cref{number}).}

\paragraph{Evaluation metrics.} 
We use two metric families: 

\noindent
\emph{Correlation with accuracy:}
The first category measures the alignment between prompt selection metrics (including our proposed metric) and the downstream accuracy of each prompt. 
This evaluation contrasts the relative quality of prompts based on their accuracy with their prompt-selection accuracy. 
Specifically, for each prompt, we compute the prompt selection metric score (MI, MI + \lmflatness, which uses only task inputs) and the prompt's accuracy on the test set. 
Given a collection of such paired numbers, we compute their correlation.  
A high correlation indicates that this prompt-selection metric can serve as a ``surrogate'' (proxy) for selecting the most accurate prompt, bypassing the direct maximization of accuracy which often demands extra held-out labeled data.


\noindent
\emph{Ranking evaluation:}
Since correlations are sensitive and brittle to outliers \cite{anscombe1973graphs}, we further use different metrics for best-performance prompt retrieval. Specifically, we use NDCG@1 \cite{jarvelin2000ir}, NDCG@3, and Rate. NDCG is a common metric for ranking quality in information retrieval. Here, we take prompts' performance as their quality score for NDCG. We denote the prompt selected by metric (e.g., highest $\operatorname{MI}$ or lowest $\sensitivity$) as $\hat{p}$, and \rate{} is defined as follows:
{
\setlength{\abovedisplayskip}{5pt}
\setlength{\belowdisplayskip}{6.5pt}
\begin{equation}\label{method::rate}
    \operatorname{\rate} = \frac{\operatorname{Performance}(\hat{p})}{\operatorname{Performance}(p_{o})},
\end{equation}}where $p_{o}$ refers to the prompt that achieves the best performance on the task.
{Intuitively, {\tt Rate} }
reflects the performance of the selected prompt compared to the best prompt, and it 
is a real-valued number between 0 and 1.
{A larger \rate{} corresponds to a better selected prompt $\hat{p}$.}

\noindent{\textbf{Flatness is complementary to \mi{} and \sensitivity.}
The correlation results are in \autoref{correlation1} (detailed numbers in \autoref{app:corr}). 
\autoref{correlation1} (first row) shows that correlations are higher for \mi+\lmflatness{} and \sensitivity+\lmflatness{} than for metrics without \lmflatness. In other words, combining existing (\mi{} or \sensitivity{})} with flatness results in a more effective prompt selection metric that correlates better with test accuracy.

We find similar results in the ranking evaluation illustrated in \autoref{fig2} (full results in \autoref{app:ret}). 
{In all benchmarks, metrics incorporating flatness generally surpass those without it, highlighting the importance of utilizing prompt flatness in the prompt selection process.}

\noindent
\textbf{Flatness \ul{alone} does \ul{not} help.}  
As shown in \autoref{correlation1}, \sensitivity+\lmflatness, \mi+\lmflatness{} or \mi{} generally outperforms \lmflatness, these results show the importance of combining prompt loss. Without prompt loss, prompt flatness on itself is insufficient to reflect prompt quality. Such results also stress the importance of combining prompt loss and flatness.

\begin{figure*}[ht]
    \centering
    \begin{tabular}{cc}
    \includegraphics[scale=0.291, trim=1.4cm 1.5cm 0cm 0cm]{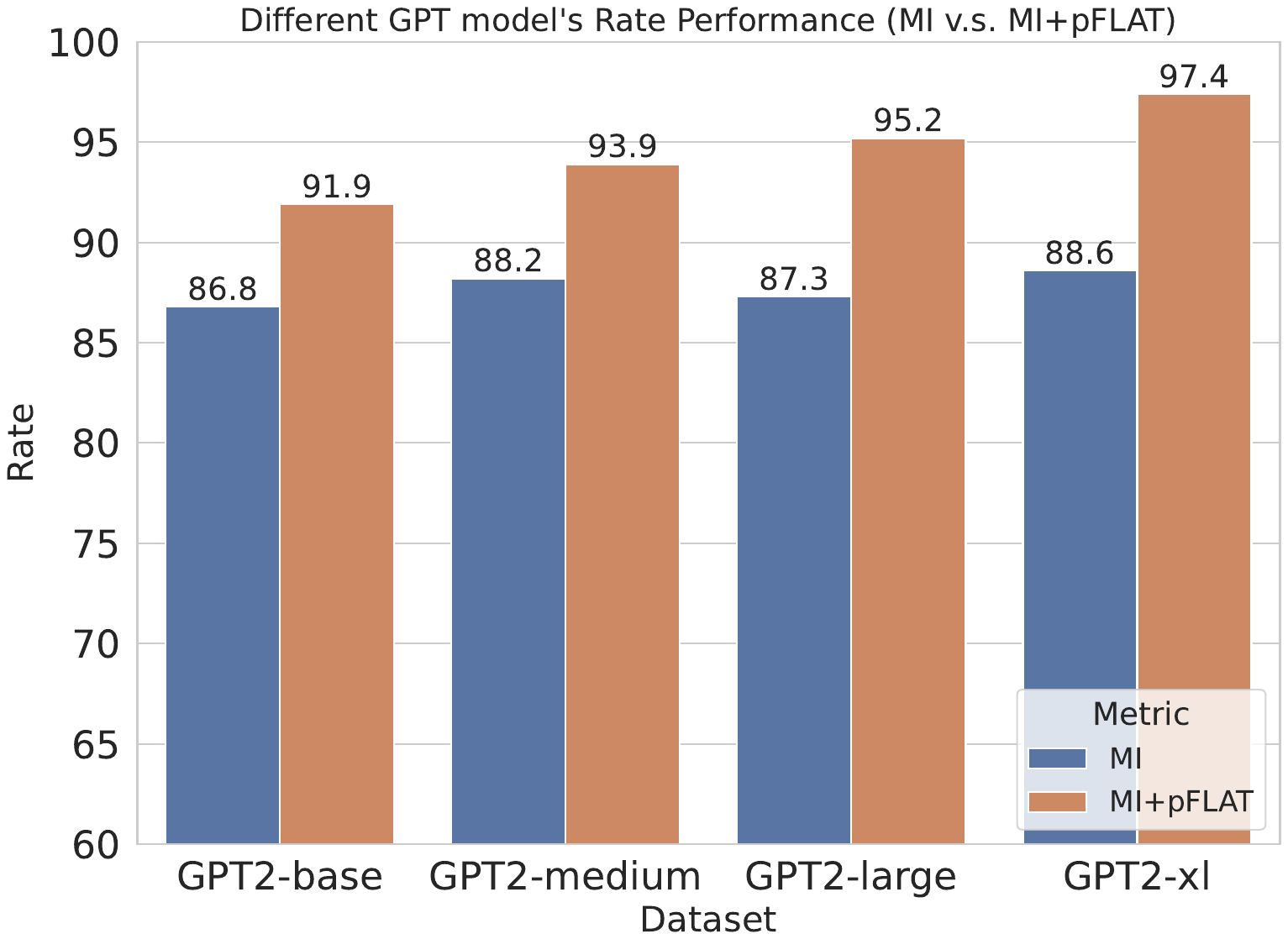}
    \includegraphics[scale=0.291, trim=1.4cm 1.5cm 0cm 0cm]{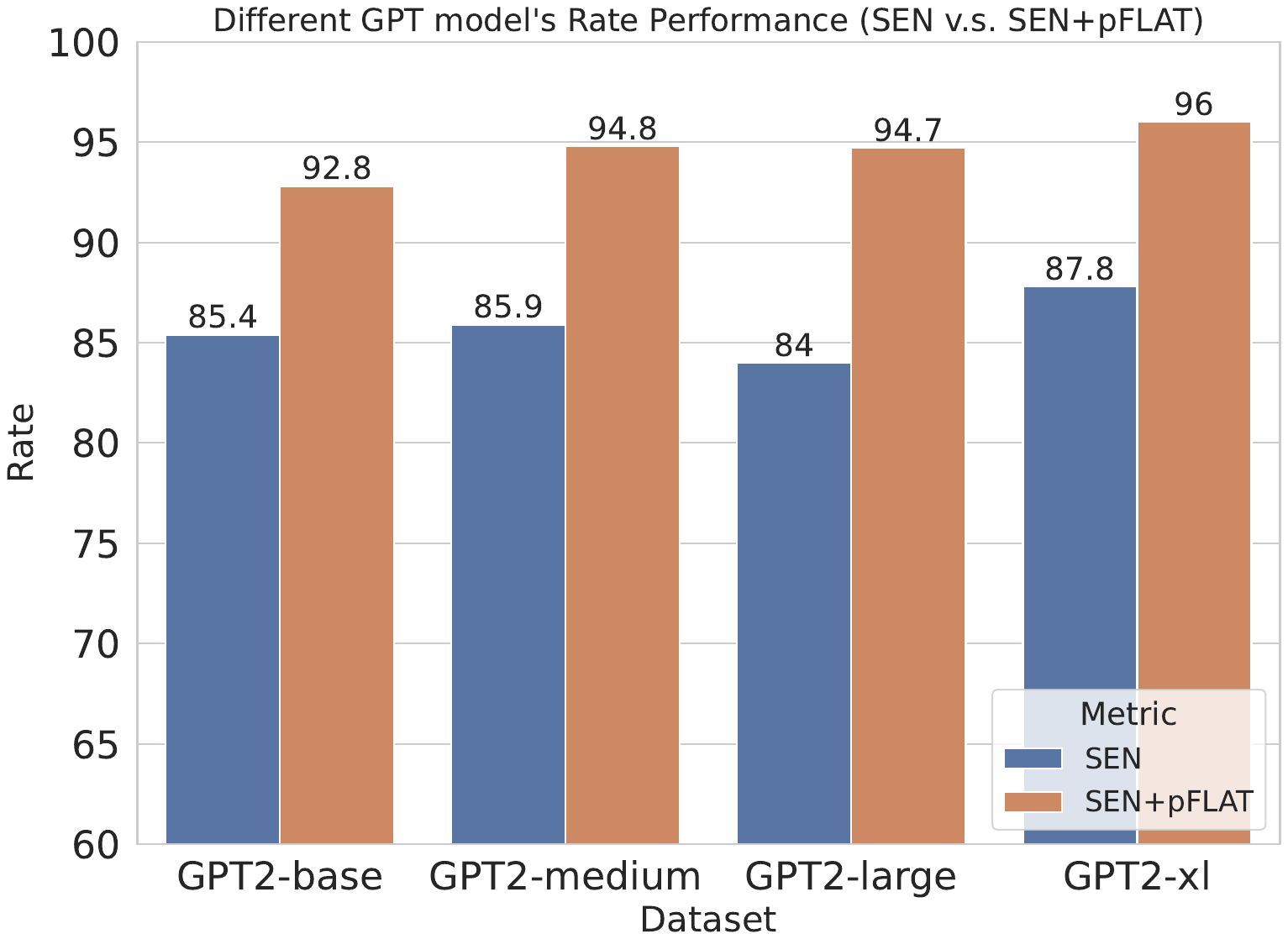}
    \end{tabular}
    \caption{\rate{} (reflecting the ability to select better prompts) evaluation computed prompt selection 
    across four model sizes, evaluated on the AGNews dataset. Combining prompt loss and flatness  (\mi+\lmflatness{} or \sensitivity+\lmflatness{}) is consistently better than MI/Sen alone across different model types. More detailed results are deferred to \autoref{app:ret}.}
    \label{ffiig}
\end{figure*}

\section{Further Analysis}\label{sec:analysis}

\subsection{Continuous Prompt Selection}
In addition to text form (discrete) prompt, we also test out the effectiveness of flatness for continuous prompt optimization (also known as `prefix-tuning'). Like the earlier result, introducing flatness to prefix-tuning also improves model performance.

\begin{table}[ht]\centering
\begin{tabular}{@{}cccc@{}}
\toprule
Method       & SST-2       & AGNews      & SNLI        \\ \midrule
w/o Flatness  & 92.5 (0.1) & 86.4 (0.2) & 72.5 (0.2) \\
w/ Flatness & \textbf{93.1} (0.1) & \textbf{87.3} (0.1) & \textbf{73.3} (0.2) \\ \bottomrule
\end{tabular}
\caption{
    Performance for prefix tuning with flatness and w/o flatness in a mean (standard-deviation) form. It is observed that \textbf{leveraging flatness in continuous prompt tuning brings improvements to performance}.
    Stronger numbers for each dataset are marked \textbf{bold}. 
}
\label{1}
\end{table}
\paragraph{Experimental setup.}
We following prefix-tuning setup of~\citet{li2021prefix} and consider three text classification
benchmarks in our experiments: SST-2~\cite{socher2013recursive},
AGNews~\cite{zhang2015character}, and SNLI~\cite{bowman2015large}. We use the GPT2-{\tt medium} as the model and set prefix length to 10 tokens for all prefix-tuning experiments. 
We train 30 epochs for SST-2 and 25 epochs for AGNews and SNLI, as suggested in \citet{yang2021robust}.

\paragraph{Implementation of flatness-aware prefix-tuning.}
To introduce flatness into prefix tuning, we leverage sharpness-aware optimizer SAM \cite{foret2020sharpness} for model optimization. We use Adam \cite{kingma2015adam} as our optimizer in the counterpart without flatness. Specifically, both cases use the same learning rate {\tt 5e-5}.

\paragraph{Results.}
As shown in \autoref{1}, prefix-tuning with flatness achieves better performance than without flatness. Such results show that \textbf{flatter continuous prompts bring better performance}, which matches our conclusions on discrete prompts.

\subsection{Influence of Model Size}
We investigate the effects of model size in our methods. 
As shown in \autoref{ffiig}, 
as the model size increases
the gap between the two metrics (e.g., MI vs MI+\lmflatness{}) measured in terms of \rate{} generally increases, indicating an increasing gain from adding \lmflatness{} to existing prompt selection for larger models.

\subsection{Impact on Sample Efficiency}
If there is enough labeled data, a reasonable approach for prompt selection is based on the accuracy of the prompt on a labeled development set (we name this baseline method ``acc''). 
Thus, a natural question concerning practicality arises: how does our method compare to prompt selection based on the accuracy of limited labeled examples? 
To perform the comparison, we select $N$ labeled data from the AGNews dataset and evaluate \rate{} (\autoref{method::rate}) for both ``acc'' baseline and our method (MI/SEN + \lmflatness{}).

\begin{figure}[!h]
    \centering
    \includegraphics[scale=0.58, trim=0.9cm 0.5cm 0cm 0cm]{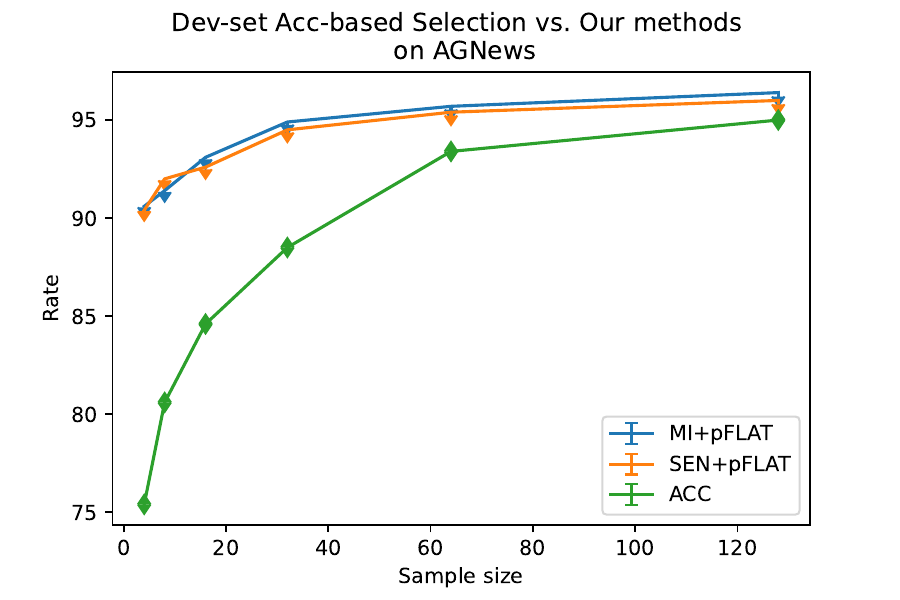}
    \caption{For development sets with varying sizes $n = 16, 32,  \cdots, 512$, the devset-acc based method (green line) selects prompts based on the accuracy of prompts on $n$ devset samples. On the other hand, our metric (\mi{}+\lmflatness{} and \sensitivity+\lmflatness{}) also use $n$ samples and achieve better performance under low-resource scenarios ($n < 500$).}
    \label{fig2222}
\end{figure}

Based on the results in \autoref{fig2222}, we observe that with little data available, our methods select a far better prompt than the ``acc'' baseline, allowing performance gains in low-data scenarios. 
This can be attributed to the fact that when the dataset is small, there may be a significant distribution shift between the development and test sets.
However, our methods, MI/Sen/\lmflatness{},
provide signals beyond labeled data and 
thus more resilient to such distribution shifts. 
Unsurprisingly, when data size grows, the gap between our method and the ``dev'' baseline decreases since the distribution shift issue is mitigated by increasing the size of the dev set. 
In conclusion, our metrics are more advantageous than development set accuracy for prompt selection in low-resource scenarios.

\begin{figure*}[ht]
    \centering
    \begin{tabular}{cc}
    (a) & (b)  \\ 
    \includegraphics[scale=0.60,trim=1.3cm 0cm 1cm 0.7cm]{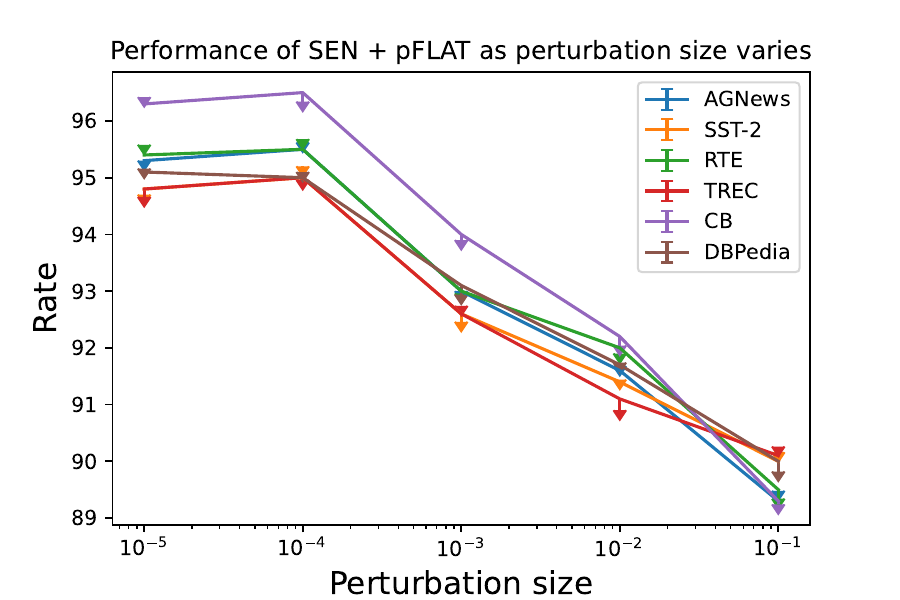} &
    \includegraphics[scale=0.60,trim=1.3cm 0cm 1cm 0.7cm]{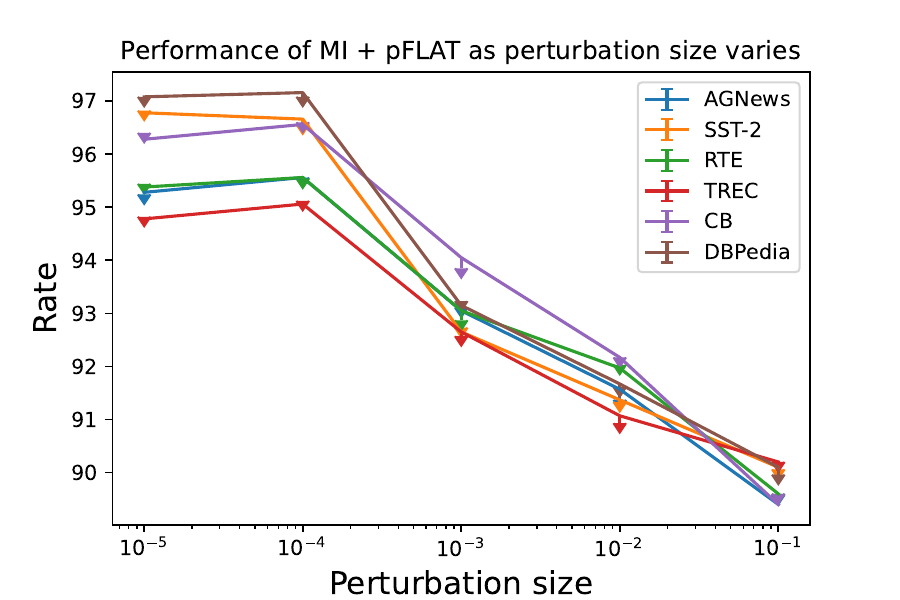}
    \end{tabular}
    \caption{(a) \rate{} of \sensitivity+\lmflatness{} as perturbation size varies. The optimal ${\epsilon}$ is around {\tt 1e-5}, as ${\epsilon}$ enlarges, the performance of \sensitivity+\lmflatness{} continues to degrade. (b) \rate{} of \mi+\lmflatness{} as perturbation size varies. The trend in \mi+\lmflatness{} is similar to (b).
    }
    \label{analysis}
\end{figure*}

\begin{figure}[!h]
    \centering
    \includegraphics[scale=0.58, trim=0.9cm 0.5cm 0cm 0cm]{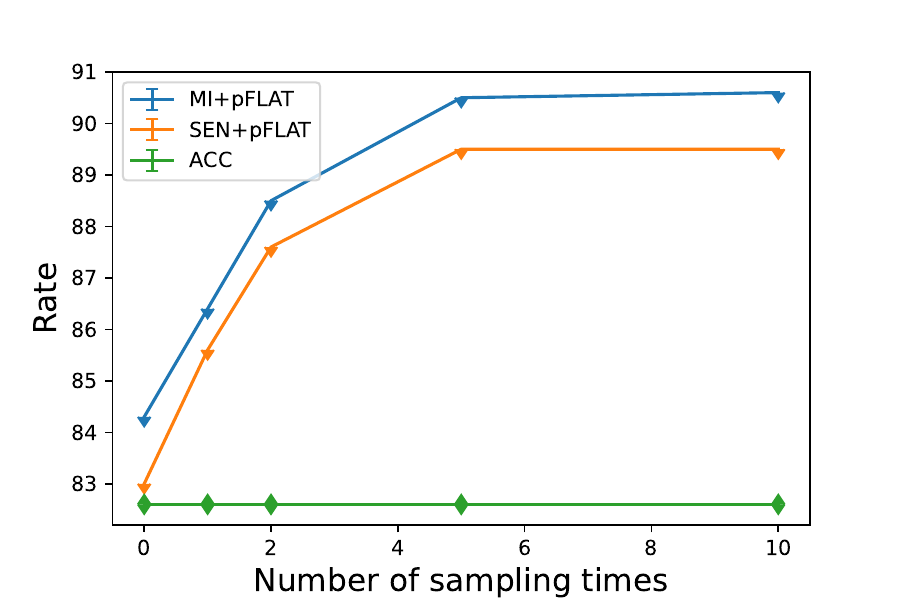}
    \caption{The trade-off between performance and sampling number $N$ in \lmflatness{}'s computation procedure. We can observe that the proper $N$ is around 5.}
    \label{fig2223}
\end{figure}

\subsection{Estimation of $\lmflatness{}$}
\label{number}
In our implementation of $\lmflatness{}$ (\autoref{all}), there are two factors that affect $\lmflatness{}$: the sampling number $N$ and the perturbation size $\sigma^2$. We explore their effects in this part.

As noted earlier, we compute prompt flatness by sampling $\epsilon$ from a standard Gaussian distribution $\mathcal{N}(0,\sigma^2)$. Since the computational cost of this estimate is proportional to the sample size $N$, the choice of $N$ is crucial for our efficiency. 
\autoref{fig2223} shows the results of an experiment showing the trade-off between $N$ and estimation quality. The results indicate that $N\approx 5$ is sufficient to provide reliable estimates for \lmflatness.

Likewise, we investigate the impact of $\sigma^2$ on the estimate. 
The results in \autoref{analysis} (a, b) indicate that the optimal perturbation size is around {\tt 1e-4}. 
When the perturbation size increases after {\tt1e-4}, the estimation error also increases.

\section{Related Work}
\paragraph{Prompt selection and engineering.}
Performance of LLMs is highly sensitive to their prompt prefix, including
the ordering of demonstrations~\cite{Lu2022FantasticallyOP} or framing of the instructions~\cite{mishra2022reframing}. 
This has motivated work prompt selection, such as the ones discussed in this work \cite{chen2022relation,sorensen2022information}. 
Beyond quantifying prompts' effectiveness, 
the literature has explored alternative ways to address LLMs' brittleness, such as chain-of-thoughts prompting~\cite{kojimalarge}, LLM self-consistency~\cite{wang2022self} and complexity~\cite{Fu2022ComplexityBasedPF}.
Our optimization-based framework does not cover these classes of prompt engineering, which we hope future work will address.

\paragraph{Algorithmic prompt generation.}
Several prior works focus on generating effective prompts to solve a given task via an LLM. Examples are RLPrompt \cite{deng2022rlprompt}, GrIPs \cite{prasad2022grips}, and Tempera \cite{zhang2023tempera}. While these works primarily focused on \emph{generating} prompts with high performance for \textit{prompt-tuning}, our goal is to \emph{identify} effective prompts for a pool of candidate prompts that is beneficial for \textit{in-context learning}. In particular, within the context of \autoref{RLPrompt}, a comparative analysis is conducted to examine the in-context learning performance of prompts generated by these approaches. The results reveal that prompts deemed suitable for fine-tuning exhibit sub-optimal performance in terms of in-context learning.

Besides, the ability to \emph{generate} prompts inevitably involves model tuning via setups like Reinforcement Learning which incurs an additional computational cost. More importantly, the quality of the generated prompts depends on the task's domain. When confronted with Out-of-Domain (OOD) tasks, these approaches tend to generate nonsensical prompts.

{
\paragraph{Continuous prompts.}
Beyond language (discrete) prompts, we show that our results also apply to continuous prompts.
In contrast to manually creating discrete prompts, one can optimize continuous prompts in embedding space, yielding better results ~\cite{lester2021power,li2021prefix,zhang2021differentiable,gu2022ppt,lang2022co,he2022hyperprompt}.
Despite higher accuracy, continuous prompt optimization is only applicable to LLMs that are publicly accessible. 
Besides, there is no evidence that continuous prompts are interpretable~\cite{khashabi2022prompt}, making it challenging to transfer insights from prompts that work well for one task to another. 
}

\paragraph{Flatness-aware language modeling}
Previous works \citep{liu2023same,mehta2021empirical} showed that flatness-aware optimization can enhance the generalization of LLM during pre-training, even if the training loss is the same. Na~et al \cite{na2022train} demonstrated that flatness-aware training increases the compression rate. Wang et al\cite{wang2022squat} showed the advantages of flatness in training encoder-only models.

{
\paragraph{Model calibration and robustness analysis.}
Model calibration focuses on adjusting LLMs' predictions to reflect human uncertainty~\cite{Holtzman2021SurfaceFC,zhao2021calibrate,jiang2022calibrating}. Calibration is related to our work as a well-calibrated LLM's confidence could be used for prompt selection.
However, calibration algorithms have remained domain/task-specific so far, restricting their applicability to the problem discussed in this paper. 
}


\section{Conclusion}
We developed a theoretical framework for prompt selection techniques that merges prompt loss and flatness, enabling the integration of previous studies to elucidate their distinctions and efficacy. Through extensive experimentation, we demonstrated the effectiveness of our proposed flatness-based metric when used in conjunction with existing ones. Our research offers valuable insights and directions for future investigations in effective prompt engineering.

\section*{Limitation}
The limitations of this study can be outlined as follows: (1) Our paper assesses the methods based on classification tasks, but they can potentially be applied to generation tasks in the future. (2) Our framework presumes that the provided collection of candidate prompts is all coherent and fluent for the intended task, despite the possibility of yielding varying results. (3) Our approach is not entirely zero-shot, since it still requires a small labeled development set for adjusting the $\alpha$ parameter.

\section*{Ethical Considerations}
To the best of our knowledge, the paper does not pose any immediate ethical concerns.

\section*{Acknowledgments}
We thank the students in the Center for Language and Speech Technologies (CLSP) for their insightful feedback.
The authors would like to thank anonymous reviewers for their constructive feedback. 
This project is supported by generous gifts from Johns Hopkins University, Allen Institute for AI, and Amazon. 
GPU machines for conducting experiments were provided by ARCH Rockfish cluster (\url{https://www.arch.jhu.edu}).

\bibliography{custom}

\providecommand{\CNFX}[1]{{\em{\textrm{(#1)}}}}
\begin{thebibliography}{51}
\expandafter\ifx\csname natexlab\endcsname\relax\def\natexlab#1{#1}\fi

\bibitem[{Andriushchenko and Flammarion(2022)}]{andriushchenko2022towards}
Maksym Andriushchenko and Nicolas Flammarion. 2022.
\newblock \href {https://arxiv.org/pdf/2206.06232v1.pdf} {Towards understanding
  sharpness-aware minimization}.
\newblock In \emph{International Conference on Machine Learning \CNFX{ICML}}.

\bibitem[{Anscombe(1973)}]{anscombe1973graphs}
Francis~J Anscombe. 1973.
\newblock \href
  {https://link.springer.com/chapter/10.1007/978-3-540-71915-1_35} {Graphs in
  statistical analysis}.
\newblock \emph{The american statistician}, 27(1):17--21.

\bibitem[{Baldassi et~al.(2020)Baldassi, Pittorino, and
  Zecchina}]{baldassi2020shaping}
Carlo Baldassi, Fabrizio Pittorino, and Riccardo Zecchina. 2020.
\newblock \href {https://www.pnas.org/doi/10.1073/pnas.1908636117} {Shaping the
  learning landscape in neural networks around wide flat minima}.
\newblock \emph{Proceedings of the National Academy of Sciences},
  117(1):161--170.

\bibitem[{Bowman et~al.(2015)Bowman, Angeli, Potts, and
  Manning}]{bowman2015large}
Samuel Bowman, Gabor Angeli, Christopher Potts, and Christopher~D Manning.
  2015.
\newblock \href {https://aclanthology.org/D15-1075/} {A large annotated corpus
  for learning natural language inference}.
\newblock In \emph{Conference on Empirical Methods in Natural Language
  Processing \CNFX{EMNLP}}.

\bibitem[{Brown et~al.(2020)Brown, Mann, Ryder, Subbiah, Kaplan, Dhariwal,
  Neelakantan, Shyam, Sastry, Askell et~al.}]{brown2020language}
Tom Brown, Benjamin Mann, Nick Ryder, Melanie Subbiah, Jared~D Kaplan, Prafulla
  Dhariwal, Arvind Neelakantan, Pranav Shyam, Girish Sastry, Amanda Askell,
  et~al. 2020.
\newblock \href {https://arxiv.org/abs/2005.14165} {Language models are
  few-shot learners}.
\newblock \emph{Advances in Neural Information Processing Systems
  \CNFX{NeurIPS}}, 33.

\bibitem[{Chen et~al.(2022)Chen, Zhao, Yu, McKeown, and He}]{chen2022relation}
Yanda Chen, Chen Zhao, Zhou Yu, Kathleen McKeown, and He~He. 2022.
\newblock \href {https://arxiv.org/abs/2209.07661} {On the relation between
  sensitivity and accuracy in in-context learning}.
\newblock \emph{arXiv preprint arXiv:2209.07661}.

\bibitem[{Dagan et~al.(2005)Dagan, Glickman, and Magnini}]{dagan2005pascal}
Ido Dagan, Oren Glickman, and Bernardo Magnini. 2005.
\newblock \href {https://link.springer.com/chapter/10.1007/11736790\_9} {The
  pascal recognising textual entailment challenge}.
\newblock In \emph{Machine Learning Challenges Workshop}.

\bibitem[{De~Marneffe et~al.(2019)De~Marneffe, Simons, and
  Tonhauser}]{de2019commitmentbank}
Marie-Catherine De~Marneffe, Mandy Simons, and Judith Tonhauser. 2019.
\newblock \href
  {https://ojs.ub.uni-konstanz.de/sub/index.php/sub/article/view/601} {The
  {CommitmentBank}: Investigating projection in naturally occurring discourse}.
\newblock In \emph{proceedings of Sinn und Bedeutung}.

\bibitem[{Deng et~al.(2022)Deng, Wang, Hsieh, Wang, Guo, Shu, Song, Xing, and
  Hu}]{deng2022rlprompt}
Mingkai Deng, Jianyu Wang, Cheng-Ping Hsieh, Yihan Wang, Han Guo, Tianmin Shu,
  Meng Song, Eric~P Xing, and Zhiting Hu. 2022.
\newblock \href {https://arxiv.org/pdf/2205.12548.pdf} {{RLPrompt:} optimizing
  discrete text prompts with reinforcement learning}.

\bibitem[{Elazar et~al.(2021)Elazar, Kassner, Ravfogel, Ravichander, Hovy,
  Sch{\"{u}}tze, and Goldberg}]{DBLP:journals/tacl/ElazarKRRHSG21}
Yanai Elazar, Nora Kassner, Shauli Ravfogel, Abhilasha Ravichander, Eduard~H.
  Hovy, Hinrich Sch{\"{u}}tze, and Yoav Goldberg. 2021.
\newblock \href {https://doi.org/10.1162/tacl\_a\_00410} {Measuring and
  improving consistency in pretrained language models}.
\newblock \emph{Trans. Assoc. Comput. Linguistics}, 9:1012--1031.

\bibitem[{Foret et~al.(2020)Foret, Kleiner, Mobahi, and
  Neyshabur}]{foret2020sharpness}
Pierre Foret, Ariel Kleiner, Hossein Mobahi, and Behnam Neyshabur. 2020.
\newblock \href {https://arxiv.org/pdf/2010.01412.pdf} {Sharpness-aware
  minimization for efficiently improving generalization}.
\newblock In \emph{International Conference on Learning Representations
  \CNFX{ICLR}}.

\bibitem[{Fu et~al.(2022)Fu, Peng, Sabharwal, Clark, and
  Khot}]{Fu2022ComplexityBasedPF}
Yao Fu, Hao-Chun Peng, Ashish Sabharwal, Peter Clark, and Tushar Khot. 2022.
\newblock \href {https://arxiv.org/pdf/2210.00720.pdf} {Complexity-based
  prompting for multi-step reasoning}.
\newblock \emph{International Conference on Learning Representations
  \CNFX{ICLR}}.

\bibitem[{Gu et~al.(2022)Gu, Han, Liu, and Huang}]{gu2022ppt}
Yuxian Gu, Xu~Han, Zhiyuan Liu, and Minlie Huang. 2022.
\newblock \href {https://arxiv.org/abs/2109.04332} {Ppt: Pre-trained prompt
  tuning for few-shot learning}.
\newblock In \emph{Annual Meeting of the Association for Computational
  Linguistics \CNFX{ACL}}.

\bibitem[{He et~al.(2022)He, Zheng, Tay, Gupta, Du, Aribandi, Zhao, Li, Chen,
  Metzler et~al.}]{he2022hyperprompt}
Yun He, Steven Zheng, Yi~Tay, Jai Gupta, Yu~Du, Vamsi Aribandi, Zhe Zhao,
  YaGuang Li, Zhao Chen, Donald Metzler, et~al. 2022.
\newblock \href {https://proceedings.mlr.press/v162/he22f/he22f.pdf}
  {{HyperPrompt:} prompt-based task-conditioning of transformers}.
\newblock In \emph{International Conference on Machine Learning \CNFX{ICML}},
  pages 8678--8690.

\bibitem[{Hochreiter and Schmidhuber(1994)}]{hochreiter1994simplifying}
Sepp Hochreiter and J{\"u}rgen Schmidhuber. 1994.
\newblock \href
  {https://proceedings.neurips.cc/paper/1994/file/01882513d5fa7c329e940dda99b12147-Paper.pdf}
  {Simplifying neural nets by discovering flat minima}.
\newblock In \emph{Advances in Neural Information Processing Systems
  \CNFX{NeurIPS}}.

\bibitem[{Holtzman et~al.(2021)Holtzman, West, Schwartz, Choi, and
  Zettlemoyer}]{Holtzman2021SurfaceFC}
Ari Holtzman, Peter West, Vered Schwartz, Yejin Choi, and Luke Zettlemoyer.
  2021.
\newblock \href {https://arxiv.org/pdf/2104.08315.pdf} {Surface form
  competition: Why the highest probability answer isn’t always right}.
\newblock \emph{Conference on Empirical Methods in Natural Language Processing
  \CNFX{EMNLP}}.

\bibitem[{J{\"a}rvelin(2000)}]{jarvelin2000ir}
Kalervo J{\"a}rvelin. 2000.
\newblock \href
  {https://dl.acm.org/doi/pdf/10.1145/3130348.3130374?casa_token=6NMyYwJm_UgAAAAA:sVO9rR-AIamAiNCvtcuZ_oRfRUhJoJFJugURBIRhEbAcOLkbiXuZihnb9KWA7Bc4jQ8zC3RHmwUsag}
  {{IR} evaluation methods for retrieving highly relevant documents.}
\newblock In \emph{Conference of the Association for Computing Machinery
  Special Interest Group in Information Retrieval \CNFX{SIGIR}}, pages 41--48.
  ACM.

\bibitem[{Jiang et~al.(2022)Jiang, Liu, and Van~Durme}]{jiang2022calibrating}
Zheng~Ping Jiang, Anqi Liu, and Benjamin Van~Durme. 2022.
\newblock \href {https://aclanthology.org/2022.emnlp-main.170/} {Calibrating
  zero-shot cross-lingual (un-) structured predictions}.
\newblock In \emph{Conference on Empirical Methods in Natural Language
  Processing \CNFX{EMNLP}}.

\bibitem[{Jiang et~al.(2020)Jiang, Xu, Araki, and
  Neubig}]{DBLP:journals/tacl/JiangXAN20}
Zhengbao Jiang, Frank~F. Xu, Jun Araki, and Graham Neubig. 2020.
\newblock \href {https://doi.org/10.1162/tacl\_a\_00324} {How can we know what
  language models know}.
\newblock \emph{Trans. Assoc. Comput. Linguistics}, 8:423--438.

\bibitem[{Keskar et~al.(2017)Keskar, Mudigere, Nocedal, Smelyanskiy, and
  Tang}]{keskar2017sharpminima}
Nitish~Shirish Keskar, Dheevatsa Mudigere, Jorge Nocedal, Mikhail Smelyanskiy,
  and Ping Tak~Peter Tang. 2017.
\newblock \href {https://arxiv.org/pdf/1609.04836.pdf,} {On large-batch
  training for deep learning: Generalization gap and sharp minima}.
\newblock In \emph{International Conference on Learning Representations
  \CNFX{ICLR}}.

\bibitem[{Khashabi et~al.(2022)Khashabi, Lyu, Min, Qin, Richardson, Welleck,
  Hajishirzi, Khot, Sabharwal, Singh, and Choi}]{khashabi2022prompt}
Daniel Khashabi, Xinxi Lyu, Sewon Min, Lianhui Qin, Kyle Richardson, Sean
  Welleck, Hannaneh Hajishirzi, Tushar Khot, Ashish Sabharwal, Sameer Singh,
  and Yejin Choi. 2022.
\newblock \href {https://arxiv.org/abs/2112.08348} {{Prompt Waywardness}: The
  curious case of discretized interpretation of continuous prompts}.
\newblock In \emph{Conference of the North American Chapter of the Association
  for Computational Linguistics \CNFX{NAACL}}.

\bibitem[{Kingma and Ba(2015)}]{kingma2015adam}
Diederik~P. Kingma and Jimmy Ba. 2015.
\newblock \href {https://arxiv.org/abs/1412.6980} {Adam: A method for
  stochastic optimization}.
\newblock abs/1412.6980.

\bibitem[{Kojima et~al.(2022)Kojima, Gu, Reid, Matsuo, and
  Iwasawa}]{kojimalarge}
Takeshi Kojima, Shixiang~Shane Gu, Machel Reid, Yutaka Matsuo, and Yusuke
  Iwasawa. 2022.
\newblock \href {https://arxiv.org/abs/2205.11916} {Large language models are
  zero-shot reasoners}.
\newblock In \emph{Advances in Neural Information Processing Systems
  \CNFX{NeurIPS}}.

\bibitem[{Lang et~al.(2022)Lang, Agrawal, Kim, and Sontag}]{lang2022co}
Hunter Lang, Monica~N Agrawal, Yoon Kim, and David Sontag. 2022.
\newblock \href {https://arxiv.org/abs/2202.00828} {Co-training improves
  prompt-based learning for large language models}.
\newblock In \emph{International Conference on Machine Learning \CNFX{ICML}}.

\bibitem[{Lester et~al.(2021)Lester, Al-Rfou, and Constant}]{lester2021power}
Brian Lester, Rami Al-Rfou, and Noah Constant. 2021.
\newblock \href {https://arxiv.org/abs/2104.08691} {The power of scale for
  parameter-efficient prompt tuning}.
\newblock In \emph{Conference on Empirical Methods in Natural Language
  Processing \CNFX{EMNLP}}.

\bibitem[{Li and Liang(2021)}]{li2021prefix}
Xiang~Lisa Li and Percy Liang. 2021.
\newblock \href {https://arxiv.org/pdf/2101.00190.pdf} {Prefix-tuning:
  Optimizing continuous prompts for generation}.
\newblock In \emph{Annual Meeting of the Association for Computational
  Linguistics \CNFX{ACL}}.

\bibitem[{Liu et~al.(2023)Liu, Xie, Li, and Ma}]{liu2023same}
Hong Liu, Sang~Michael Xie, Zhiyuan Li, and Tengyu Ma. 2023.
\newblock \href {https://proceedings.mlr.press/v202/liu23ao.html} {Same
  pre-training loss, better downstream: Implicit bias matters for language
  models}.
\newblock In \emph{International Conference on Machine Learning}, pages
  22188--22214. PMLR.

\bibitem[{Lu et~al.(2022)Lu, Bartolo, Moore, Riedel, and
  Stenetorp}]{Lu2022FantasticallyOP}
Yao Lu, Max Bartolo, Alastair Moore, Sebastian Riedel, and Pontus Stenetorp.
  2022.
\newblock \href {https://arxiv.org/pdf/2104.08786.pdf} {Fantastically ordered
  prompts and where to find them: Overcoming few-shot prompt order
  sensitivity}.
\newblock In \emph{Annual Meeting of the Association for Computational
  Linguistics \CNFX{ACL}}.

\bibitem[{Mehta et~al.(2021)Mehta, Patil, Chandar, and
  Strubell}]{mehta2021empirical}
Sanket~Vaibhav Mehta, Darshan Patil, Sarath Chandar, and Emma Strubell. 2021.
\newblock \href {https://www.cs.cmu.edu/~svmehta/papers/pretrainlll.pdf} {An
  empirical investigation of the role of pre-training in lifelong learning}.
\newblock \emph{arXiv preprint arXiv:2112.09153}.

\bibitem[{Mishra et~al.(2022)Mishra, Khashabi, Baral, Choi, and
  Hajishirzi}]{mishra2022reframing}
Swaroop Mishra, Daniel Khashabi, Chitta Baral, Yejin Choi, and Hannaneh
  Hajishirzi. 2022.
\newblock \href {https://arxiv.org/abs/2109.07830} {Reframing instructional
  prompts to gptk's language}.
\newblock In \emph{Annual Meeting of the Association for Computational
  Linguistics \CNFX{ACL} - Findings}.

\bibitem[{Na et~al.(2022)Na, Mehta, and Strubell}]{na2022train}
Clara Na, Sanket~Vaibhav Mehta, and Emma Strubell. 2022.
\newblock \href {https://aclanthology.org/2022.findings-emnlp.361.pdf} {Train
  flat, then compress: Sharpness-aware minimization learns more compressible
  models}.
\newblock In \emph{Findings of the Association for Computational Linguistics:
  EMNLP 2022}, pages 4909--4936.

\bibitem[{Perez et~al.(2021)Perez, Kiela, and Cho}]{perez2021true}
Ethan Perez, Douwe Kiela, and Kyunghyun Cho. 2021.
\newblock \href
  {https://proceedings.neurips.cc/paper/2021/file/5c04925674920eb58467fb52ce4ef728-Paper.pdf}
  {True few-shot learning with language models}.
\newblock In \emph{Advances in Neural Information Processing Systems
  \CNFX{NeurIPS}}.

\bibitem[{Prasad et~al.(2023)Prasad, Hase, Zhou, and Bansal}]{prasad2022grips}
Archiki Prasad, Peter Hase, Xiang Zhou, and Mohit Bansal. 2023.
\newblock \href {https://arxiv.org/pdf/2203.07281.pdf} {Grips: Gradient-free,
  edit-based instruction search for prompting large language models}.
\newblock \emph{Conference of the European Chapter of the Association for
  Computational Linguistics \CNFX{EACL}}.

\bibitem[{Radford et~al.(2019)Radford, Wu, Child, Luan, Amodei, Sutskever
  et~al.}]{radford2019language}
Alec Radford, Jeffrey Wu, Rewon Child, David Luan, Dario Amodei, Ilya
  Sutskever, et~al. 2019.
\newblock \href {https://openai.com/blog/better-language-models/} {Language
  models are unsupervised multitask learners}.
\newblock \emph{OpenAI blog}.

\bibitem[{Reynolds and McDonell(2021)}]{reynolds2021prompt}
Laria Reynolds and Kyle McDonell. 2021.
\newblock \href
  {https://dl.acm.org/doi/pdf/10.1145/3411763.3451760?casa_token=ijNC_4bqaaAAAAAA:bmT5di0BHvPiIv4k3_cw6S2YkKPUhAYeqKVsqgY7KwSTLR9DG7Jsa04ap8oGk3XM-BAIr8QXjh0KPA}
  {Prompt programming for large language models: Beyond the few-shot paradigm}.
\newblock In \emph{Conference on Human Factors in Computing Systems
  \CNFX{CHI}}.

\bibitem[{Schick and Sch{\"u}tze(2021{\natexlab{a}})}]{schick2021exploiting}
Timo Schick and Hinrich Sch{\"u}tze. 2021{\natexlab{a}}.
\newblock \href {https://arxiv.org/pdf/2001.07676.pdf} {Exploiting
  cloze-questions for few-shot text classification and natural language
  inference}.
\newblock In \emph{Conference of the European Chapter of the Association for
  Computational Linguistics \CNFX{EACL}}, pages 255--269.

\bibitem[{Schick and Sch{\"u}tze(2021{\natexlab{b}})}]{Schick2020FewShotTG}
Timo Schick and Hinrich Sch{\"u}tze. 2021{\natexlab{b}}.
\newblock \href {https://arxiv.org/pdf/2012.11926.pdf} {Few-shot text
  generation with pattern-exploiting training}.
\newblock \emph{Conference on Empirical Methods in Natural Language Processing
  \CNFX{EMNLP}}.

\bibitem[{Socher et~al.(2013)Socher, Perelygin, Wu, Chuang, Manning, Ng, and
  Potts}]{socher2013recursive}
Richard Socher, Alex Perelygin, Jean Wu, Jason Chuang, Christopher~D Manning,
  Andrew~Y Ng, and Christopher Potts. 2013.
\newblock \href {https://aclanthology.org/D13-1170.pdf} {Recursive deep models
  for semantic compositionality over a sentiment treebank}.
\newblock In \emph{Conference on Empirical Methods in Natural Language
  Processing \CNFX{EMNLP}}, pages 1631--1642.

\bibitem[{Sorensen et~al.(2022)Sorensen, Robinson, Rytting, Shaw, Rogers,
  Delorey, Khalil, Fulda, and Wingate}]{sorensen2022information}
Taylor Sorensen, Joshua Robinson, Christopher Rytting, Alexander Shaw, Kyle
  Rogers, Alexia Delorey, Mahmoud Khalil, Nancy Fulda, and David Wingate. 2022.
\newblock \href {https://arxiv.org/abs/2203.11364} {An information-theoretic
  approach to prompt engineering without ground truth labels}.
\newblock In \emph{Annual Meeting of the Association for Computational
  Linguistics \CNFX{ACL}}, pages 819--862.

\bibitem[{Stutz et~al.(2021)Stutz, Hein, and Schiele}]{stutz2021relating}
David Stutz, Matthias Hein, and Bernt Schiele. 2021.
\newblock \href
  {http://openaccess.thecvf.com/content/ICCV2021/papers/Stutz_Relating_Adversarially_Robust_Generalization_to_Flat_Minima_ICCV_2021_paper.pdf}
  {Relating adversarially robust generalization to flat minima}.
\newblock In \emph{IEEE Conference on Computer Vision and Pattern Recognition
  \CNFX{CVPR}}, pages 7807--7817.

\bibitem[{Voorhees and Tice(2000)}]{voorhees2000building}
Ellen~M Voorhees and Dawn~M Tice. 2000.
\newblock \href {https://dl.acm.org/doi/pdf/10.1145/345508.345577} {Building a
  question answering test collection}.
\newblock In \emph{Conference of the Association for Computing Machinery
  Special Interest Group in Information Retrieval \CNFX{SIGIR}}.

\bibitem[{Wang et~al.(2022{\natexlab{a}})Wang, Wei, Schuurmans, Le, Chi, and
  Zhou}]{wang2022self}
Xuezhi Wang, Jason Wei, Dale Schuurmans, Quoc Le, Ed~Chi, and Denny Zhou.
  2022{\natexlab{a}}.
\newblock \href {https://arxiv.org/abs/2203.11171} {Self-consistency improves
  chain of thought reasoning in language models}.
\newblock \emph{arXiv preprint arXiv:2203.11171}.

\bibitem[{Wang et~al.(2022{\natexlab{b}})Wang, Li, Qu, Metze, and
  Strubell}]{wang2022squat}
Zheng Wang, Juncheng~B Li, Shuhui Qu, Florian Metze, and Emma Strubell.
  2022{\natexlab{b}}.
\newblock \href {https://arxiv.org/pdf/2210.07171.pdf} {Squat: Sharpness-and
  quantization-aware training for bert}.
\newblock \emph{arXiv preprint arXiv:2210.07171}.

\bibitem[{Yang and Liu(2021)}]{yang2021robust}
Zonghan Yang and Yang Liu. 2021.
\newblock \href {https://minicheshire.github.io/papers/ICLR22_slides.pdf} {On
  robust prefix-tuning for text classification}.
\newblock In \emph{International Conference on Learning Representations
  \CNFX{ICLR}}.

\bibitem[{Zhang et~al.(2022)Zhang, Li, Chen, Deng, Bi, Tan, Huang, and
  Chen}]{zhang2021differentiable}
Ningyu Zhang, Luoqiu Li, Xiang Chen, Shumin Deng, Zhen Bi, Chuanqi Tan, Fei
  Huang, and Huajun Chen. 2022.
\newblock \href {https://arxiv.org/pdf/2108.13161.pdf} {Differentiable prompt
  makes pre-trained language models better few-shot learners}.
\newblock \emph{International Conference on Learning Representations
  \CNFX{ICLR}}.

\bibitem[{Zhang et~al.(2023{\natexlab{a}})Zhang, Wang, Zhou, Schuurmans, and
  Gonzalez}]{zhang2023tempera}
Tianjun Zhang, Xuezhi Wang, Denny Zhou, Dale Schuurmans, and Joseph~E Gonzalez.
  2023{\natexlab{a}}.
\newblock \href {https://openreview.net/pdf?id=gSHyqBijPFO} {{TEMPERA}:
  Test-time prompt editing via reinforcement learning}.
\newblock In \emph{International Conference on Learning Representations
  \CNFX{ICLR}}.

\bibitem[{Zhang et~al.(2015)Zhang, Zhao, and LeCun}]{zhang2015character}
Xiang Zhang, Junbo Zhao, and Yann LeCun. 2015.
\newblock \href
  {https://proceedings.neurips.cc/paper/2015/file/250cf8b51c773f3f8dc8b4be867a9a02-Paper.pdf}
  {Character-level convolutional networks for text classification}.
\newblock In \emph{Advances in Neural Information Processing Systems
  \CNFX{NeurIPS}}.

\bibitem[{Zhang et~al.(2023{\natexlab{b}})Zhang, Xu, Yu, Zou, and
  Cui}]{zhang2023gradient}
Xingxuan Zhang, Renzhe Xu, Han Yu, Hao Zou, and Peng Cui. 2023{\natexlab{b}}.
\newblock \href {https://openreview.net/forum?id=z4eslwuymzQ} {Gradient norm
  regularizer seeks flat minima and improves generalization}.

\bibitem[{Zhao et~al.(2022)Zhao, Zhang, and
  Hu}]{penalizinggradientnorm2022zhao}
Yang Zhao, Hao Zhang, and Xiuyuan Hu. 2022.
\newblock \href {https://proceedings.mlr.press/v162/zhao22i.html} {Penalizing
  gradient norm for efficiently improving generalization in deep learning}.
\newblock In \emph{International Conference on Machine Learning \CNFX{ICML}}.

\bibitem[{Zhao et~al.(2021)Zhao, Wallace, Feng, Klein, and
  Singh}]{zhao2021calibrate}
Zihao Zhao, Eric Wallace, Shi Feng, Dan Klein, and Sameer Singh. 2021.
\newblock \href {http://proceedings.mlr.press/v139/zhao21c/zhao21c.pdf}
  {Calibrate before use: Improving few-shot performance of language models}.
\newblock In \emph{International Conference on Machine Learning \CNFX{ICML}},
  pages 12697--12706.

\bibitem[{Zheng et~al.(2021)Zheng, Zhang, and Mao}]{zheng2021regularizing}
Yaowei Zheng, Richong Zhang, and Yongyi Mao. 2021.
\newblock \href
  {https://openaccess.thecvf.com/content/CVPR2021/papers/Zheng_Regularizing_Neural_Networks_via_Adversarial_Model_Perturbation_CVPR_2021_paper.pdf}
  {Regularizing neural networks via adversarial model perturbation}.
\newblock In \emph{IEEE Conference on Computer Vision and Pattern Recognition
  \CNFX{CVPR}}, pages 8156--8165.

\end{thebibliography}
\bibliographystyle{acl_natbib}

\clearpage

\onecolumn

\appendix
\section*{\LARGE{Supplementary Material}}

\begin{table}[ht]
    \centering
    \footnotesize
    \begin{tabular}{cl}
    Appendix     & Contents  \\ \toprule
    \autoref{app:prompt-loss}     &  \begin{tabular}[c]{@{}l@{}} Formal connection between optimizing 
    Sen/MI  and  are minimization of surrogates  on prompt loss\end{tabular} \\ \midrule
    \autoref{app:prompt-flatness} & \begin{tabular}[c]{@{}l@{}}Formal connection between \lmflatness{}  and true flatness\end{tabular} \\ \midrule
    \autoref{app:corr} & \begin{tabular}[c]{@{}l@{}}Extra results on the  correlation evaluation\end{tabular} \\ \midrule
    \autoref{app:ret} & \begin{tabular}[c]{@{}l@{}}Extra results on the retrieval evaluation\end{tabular} \\ \midrule
    \autoref{RLPrompt} & \begin{tabular}[c]{@{}l@{}}ICL performance comparison between prompt selected by our method and other prompt generation methods.\end{tabular} \\ \midrule
    \autoref{app:discrete-prompt} & \begin{tabular}[c]{@{}l@{}}The instruction set used in our paper on each benchmark.\end{tabular} \\ \bottomrule
\end{tabular}    
\end{table}

\section{Sen and MI are approximations (surrogates) of prompt loss}\label{app:prompt-loss}

In this section, we demonstrate that Sen \cite{sorensen2022information} and MI \cite{chen2022relation} of prompt $p$ on dataset $\dataset$ are essentially surrogates for the prompt loss $\mathcal{L}$ on $\dataset$. 

\paragraph{Mutual Information}
\citet{sorensen2022information} hypothesizes that a prompt $p_i$ with higher mutual information (MI) will align a language model to a task better. In prompt selection, MI select the prompt $\hat{p}=\operatorname{argmax}_p\left\{I\left(f_{\theta}(\dataset \circ p) ; \mathcal{Y}\right)\right\}$ and MI can be estimated as:
\begin{equation}\label{MI}
\begin{aligned}
MI(\dataset,p;\theta)&=I\left(f_{\theta}(\dataset,p);\mathcal{Y}\right)\\
        &=H(\mathcal{Y})-H\left( f_{\theta}(\dataset,p)\right)
\end{aligned}
\end{equation}
where $H$ refers to entropy, and each term is estimated in expectation using N draws $x \sim \datasetX$:
\begin{gather}
H(\mathcal{Y}) = H\left(\frac{1}{N} \sum_{x \in \datasetX} f_{\theta}(x\circ p)\right) 
\\
H\left(\mathcal{Y} \mid f_{\theta}(\dataset,p)\right) = \frac{1}{N}\sum_{x \in \datasetX} H(f_{\theta}(x\circ p))\nonumber
\end{gather}

According to the Weak Law of Large Numbers (and assume that test samples $x \in \dataset$ are independently drawn from an unknown distribution $P_{\dataset}$), it is easy to obtain that
\begin{equation}\label{abc}
\lim _{|\dataset| \rightarrow \infty} \mathbb{P}\left(\left|H(\mathcal{Y}) -H(\mathbf{E})\right| \geq \epsilon\right)=0
\end{equation}
Where $\mathbf{E}$ refers to the expectation $\mathbb{E}_{x \in P_{\dataset}}P(Y|x\circ p;\theta)$, and it is a fixed distribution once $f$ and $P_{\dataset}$ are determined. As shown by \autoref{abc}, $H(\mathcal{Y})$ converges to a constant as the test sample number increases. Now we focus on the second term of $MI(\dataset,p)$, as shown in \autoref{MI}. We also re-write it as follows:
\begin{align}
\footnotesize
&H\left(\mathcal{Y} \mid f_{\theta}(\dataset,p)\right) = \frac{1}{|\dataset|}\sum _{x \in \datasetX} H(f_{\theta}(x\circ p))\nonumber \\ \nonumber
&=\frac{1}{|\dataset|}\sum _{x \in \datasetX} [\ell_{ce}(y,f_{\theta}(x\circ p)) \\&- \operatorname{KL}(f_{\theta}(x\circ p)||y)]\nonumber\\ &=\mathcal{L}(p,\dataset, \theta) - \frac{1}{|\dataset|}\sum _{x \in \datasetX} \operatorname{KL}(f_{\theta}(x\circ p)||y) \nonumber  
\end{align}
where $\ell_{ce}$ refers to cross-entropy and $\operatorname{KL}$ is the KL divergence. The equation above illustrates a relation between the second term of $MI$ and prompt loss. The gap between $\operatorname{MI}$ and prompt loss is the average $\operatorname{KL}$ divergence. Overall, $\operatorname{MI}$ can be formulated as follows:
\begin{gather}\label{A}
        \operatorname{MI}(\dataset,p) = H(\mathbf{E}) - \mathcal{L}(p,\dataset, \theta) \\+ \frac{1}{|\dataset|}\sum _{x \in \datasetX} \operatorname{KL}(f_{\theta}(x\circ p)||y)
\end{gather}
\autoref{A} shows that maximizing mutual information is equivalent to minimizing prompt loss to a certain degree, indicating that $\operatorname{MI}$ serves as a surrogate for prompt loss.

\paragraph{Sensitivity}
Sensitivity (Sen) reflects how much the model output changes given small perturbations of the input. Sen first creates a perturbed prompt set $\mathcal{P}$ given a prompt $p$, by changing demo order $\sigma$ and adding perturbation $\epsilon$ to the prompt instruction $I$. We direct readers to the original paper \cite{chen2022relation} for details of how such prompt sets can be created. Sensitivity on one single test sample $x$ is formally denoted as follows:
\begin{gather}\footnotesize
    \sensitivity(x,p)  =  \sum_{p^{\prime} \in \mathcal{P}} \mathbf{1}\left[f_{\theta}(x\circ p) \neq f_{\theta}(x\circ p^{\prime})\right]
\end{gather}
Naturally, we can extend this sample-level metric to the dataset level. Given test samples $\dataset$, the $\sensitivity$ of prompt $p$ is defined as follows:
\begin{equation}
    \sensitivity(\dataset,p) \\= \frac{1}{N}\sum_{x \in \datasetX} \sensitivity(x,p)
\end{equation}

We can re-write the formula for Sen as follows:
\begin{equation}
\begin{aligned}
&\sensitivity(\dataset,p) = \frac{1}{N}\sum_{x \in \datasetX} \sensitivity(x,p) \\
&=\frac{1}{|\dataset|}\sum _{x \in \datasetX} \mathbf{E}_{p^{\prime}}\ell_{01}(f_{\theta}(x\circ p^{\prime}),f_{\theta}(x\circ p)) \\ 
&=\mathcal{L}(p,\dataset, \theta) - \frac{1}{|\dataset|}\sum _{x \in \datasetX} \mathbf{E}_{p^{\prime}}\ell_{01}(f_{\theta}(x \circ p^{\prime}),y) \nonumber
\end{aligned}    
\end{equation}
Note that $\mathcal{L}(p,\dataset)$ is a 0-1 loss instead of cross-entropy loss as shown in MI's derivation. The equation above shows that $\sensitivity$ can be regarded as a surrogate for the prompt loss $\mathcal{L}$. Therefore, minimizing $\sensitivity$ is partially equal to minimizing prompt loss, explaining why a low-sensitivity prompt achieves better performance, as empirically verified by \citet{chen2022relation}.

Generally, the gap between prompt loss $\mathcal{L}$ and two surrogates ($\operatorname{MI}$ and $\sensitivity$) is determined by the distance (i.e., KL divergence) between the model's prediction $f_{\theta}(x\circ p)$ distribution and ground-truth label.
When $f_{\theta}(x)$ is identical to round-truth label, $\operatorname{MI}$ and $\sensitivity$ become perfect surrogates for prompt loss $\mathcal{L}$.

\section{On the approximation gap of flatness and $\mathcal{F}$}\label{app:prompt-flatness}
This section details the approximation gap of flatness $\lmflatness{}$ towards $\mathcal{F}$. Firstly, we recall the definition of $\lmflatness{}$ and $\mathcal{F}$.

\begin{equation}\small\label{hhh1}
\begin{aligned}
\lmflatness{}(p,\dataset, \theta) &= \frac{1}{|\dataset|}\sum _{x \in \datasetX}  \mathbf{E}_{\epsilon} [\ell(f_{\theta}(p\circ x),f_{\theta+\epsilon_{1}}(p\circ x) ) \\&-\ell(f_{\theta}(p\circ x),f_{\theta+\epsilon_{2}}(p\circ x) )]
\\&=\frac{1}{|\dataset|}\sum _{x \in \datasetX}  \mathbf{E}_{\epsilon} \norm{\ell(f_{\theta}(p\circ x),f_{\theta+\epsilon}(p\circ x)}_2
\end{aligned}
\end{equation}
Also, we re-write $\mathcal{F}(p, \dataset, \theta)$ as follows:
\begin{gather}\label{hhh2}
    \mathcal{F}(p, \dataset, \theta) = \norm{
\nabla_{\theta} 
\mathcal{L}(p, \dataset, \theta)
}_2 \\
=\frac{1}{|\dataset|}\sum _{x,y \in \dataset}  \nabla_{\theta}\norm{\ell(f_{\theta}(p\circ x),y}_2
\end{gather}
Thus, the approximation gap can be obtained through \autoref{hhh1} and \autoref{hhh2}. When the model's confidence is identical to ground-truth labels, \lmflatness{} is a precise approximator of $\mathcal{F}$.

\newpage

\section{Results on correlation}\label{app:corr}
Here are the full results of correlation comparisons in our paper, as shown in \autoref{correlation}.
\begin{table*}[ht]\centering\small
\begin{tabular}{c|c|cc|cc|cc|cc|cc}
\toprule[2pt]
\multirow{2}{*}{Model}       & \multirow{2}{*}{Methods} & \multicolumn{2}{c|}{AGNews}       & \multicolumn{2}{c|}{CB}          & \multicolumn{2}{c|}{DBpedia}       & \multicolumn{2}{c|}{SST-2}         & \multicolumn{2}{c}{RTE}           \\ \cmidrule(l){3-12} 
                             &                          & \multicolumn{1}{c|}{Pr}    & Spr  & \multicolumn{1}{c|}{Pr}   & Spr  & \multicolumn{1}{c|}{Pr}    & Spr   & \multicolumn{1}{c|}{Pr}   & Spr  & \multicolumn{1}{c|}{Pr}   & Spr   \\ \midrule[2pt]
\multirow{6}{*}{GPT2-base}   & MI                       & \multicolumn{1}{c|}{21.9}  & 22.5 & \multicolumn{1}{c|}{3.5}  & 4.5  & \multicolumn{1}{c|}{\color[HTML]{FE0000}30.1}  & 25.1  & \multicolumn{1}{c|}{19.2} & 16.8 & \multicolumn{1}{c|}{20.6} & 23.5  \\
                             & Sen                      & \multicolumn{1}{c|}{8.6}   & 7.6  & \multicolumn{1}{c|}{14.3} & 13.4 & \multicolumn{1}{c|}{-10.6} & -14.2 & \multicolumn{1}{c|}{5.6}  & 3.4  & \multicolumn{1}{c|}{-9.9} & -13.9 \\
                             & \lmflatness{}                     & \multicolumn{1}{c|}{21.4}  & 19.8 & \multicolumn{1}{c|}{-9.1} & -8.1 & \multicolumn{1}{c|}{21.3}  & 22.0  & \multicolumn{1}{c|}{18.3} & 20.2 & \multicolumn{1}{c|}{20.2} & 20.4  \\
                             & MI+Sen                   & \multicolumn{1}{c|}{22.0}  & 16.4 & \multicolumn{1}{c|}{17.1} & 18.1 & \multicolumn{1}{c|}{26.0}  & 23.7  & \multicolumn{1}{c|}{20.1} & 21.2 & \multicolumn{1}{c|}{24.9} & 23.8  \\
                             & MI+\lmflatness{}                  & \multicolumn{1}{c|}{26.3}  & 25.7 & \multicolumn{1}{c|}{20.5} & 20.7 & \multicolumn{1}{c|}{24.1}  & 25.6  & \multicolumn{1}{c|}{23.4} & 21.7 & \multicolumn{1}{c|}{15.4} & 17.5  \\
                             & Sen+\lmflatness{}                 & \multicolumn{1}{c|}{\color[HTML]{FE0000}28.6}  & \color[HTML]{FE0000}29.3 & \multicolumn{1}{c|}{\color[HTML]{FE0000}28.1} & \color[HTML]{FE0000}26.7 & \multicolumn{1}{c|}{29.4}  & \color[HTML]{FE0000}28.1  & \multicolumn{1}{c|}{\color[HTML]{FE0000}23.6} & \color[HTML]{FE0000}24.5 & \multicolumn{1}{c|}{\color[HTML]{FE0000}27.2} &\color[HTML]{FE0000} 25.3  \\ \midrule
\multirow{6}{*}{GPT2-medium} & MI                       & \multicolumn{1}{c|}{27.5}  & 26.4 & \multicolumn{1}{c|}{\color[HTML]{FE0000}26.7} & 23.0 & \multicolumn{1}{c|}{28.9}  & 26.9  & \multicolumn{1}{c|}{27.1} & 25.0 & \multicolumn{1}{c|}{22.5} & 16.2  \\
                             & Sen                      & \multicolumn{1}{c|}{-11.2} & 3.5  & \multicolumn{1}{c|}{-4.5} & -7.7 & \multicolumn{1}{c|}{-8.6}  & -10.8 & \multicolumn{1}{c|}{10.1} & 5.4  & \multicolumn{1}{c|}{-8.4} & -10.3 \\
                             & \lmflatness{}                     & \multicolumn{1}{c|}{23.8}  & 26.0 & \multicolumn{1}{c|}{21.6} & 22.5 & \multicolumn{1}{c|}{20.6}  & 23.4  & \multicolumn{1}{c|}{23.8} & 23.3 & \multicolumn{1}{c|}{20.2} & 17.7  \\
                             & MI+Sen                   & \multicolumn{1}{c|}{24.7}  & 22.8 & \multicolumn{1}{c|}{18.7} & 20.4 & \multicolumn{1}{c|}{23.7}  & 27.0  & \multicolumn{1}{c|}{27.7} & 26.5 & \multicolumn{1}{c|}{11.2} & 13.0  \\
                             & MI+\lmflatness{}                  & \multicolumn{1}{c|}{\color[HTML]{FE0000}29.0}  & \color[HTML]{FE0000}30.1 & \multicolumn{1}{c|}{22.9} & 20.5 & \multicolumn{1}{c|}{29.9}  & \color[HTML]{FE0000}31.9  & \multicolumn{1}{c|}{27.0} & 28.7 & \multicolumn{1}{c|}{\color[HTML]{FE0000}25.7} &\color[HTML]{FE0000} 20.6  \\
                             & Sen+\lmflatness{}                 & \multicolumn{1}{c|}{28.1}  & 29.0 & \multicolumn{1}{c|}{23.6} & \color[HTML]{FE0000}26.6 & \multicolumn{1}{c|}{\color[HTML]{FE0000}33.1}  & 31.8  & \multicolumn{1}{c|}{32.3} & \color[HTML]{FE0000}32.7 & \multicolumn{1}{c|}{24.0} & 17.3  \\ \midrule
\multirow{6}{*}{GPT2-large}  & MI                       & \multicolumn{1}{c|}{23.4}  & 21.0 & \multicolumn{1}{c|}{20.9} & 20.9 & \multicolumn{1}{c|}{24.2}  & 27.2  & \multicolumn{1}{c|}{20.2} & 21.3 & \multicolumn{1}{c|}{22.8} & 18.6  \\
                             & Sen                      & \multicolumn{1}{c|}{11.0}  & 5.6  & \multicolumn{1}{c|}{-6.0} & -4.3 & \multicolumn{1}{c|}{-5.9}  & -8.9  & \multicolumn{1}{c|}{-6.7} & 5.8  & \multicolumn{1}{c|}{11.0} & 22.1  \\
                             & \lmflatness{}                     & \multicolumn{1}{c|}{20.0}  & 20.7 & \multicolumn{1}{c|}{19.4} & 22.4 & \multicolumn{1}{c|}{21.7}  & 22.5  & \multicolumn{1}{c|}{20.1} & 18.3 & \multicolumn{1}{c|}{22.1} & 20.2  \\
                             & MI+Sen                   & \multicolumn{1}{c|}{25.0}  & 21.3 & \multicolumn{1}{c|}{24.1} & 24.6 & \multicolumn{1}{c|}{23.0}  & 27.0  & \multicolumn{1}{c|}{25.3} & 24.0 & \multicolumn{1}{c|}{19.4} & 21.5  \\
                             & MI+\lmflatness{}                  & \multicolumn{1}{c|}{25.4}  & 26.7 & \multicolumn{1}{c|}{25.4} & \color[HTML]{FE0000}29.3 & \multicolumn{1}{c|}{\color[HTML]{FE0000}26.3}  & 27.6  & \multicolumn{1}{c|}{30.1} & 28.9 & \multicolumn{1}{c|}{\color[HTML]{FE0000}35.3} & \color[HTML]{FE0000}30.9  \\
                             & Sen+\lmflatness{}                 & \multicolumn{1}{c|}{\color[HTML]{FE0000}29.5}  & \color[HTML]{FE0000}28.8 & \multicolumn{1}{c|}{\color[HTML]{FE0000}28.0} & 28.5 & \multicolumn{1}{c|}{24.9}  & \color[HTML]{FE0000}28.4  & \multicolumn{1}{c|}{\color[HTML]{FE0000}31.3} & \color[HTML]{FE0000}30.4 & \multicolumn{1}{c|}{19.3} & 20.9  \\ \midrule
\multirow{6}{*}{GPT2-xl}     & MI                       & \multicolumn{1}{c|}{22.7}  & \color[HTML]{FE0000}24.3 & \multicolumn{1}{c|}{18.9} & 20.1 & \multicolumn{1}{c|}{24.7}  & 22.0  & \multicolumn{1}{c|}{15.3} & 16.8 & \multicolumn{1}{c|}{14.7} & 21.2  \\
                             & Sen                      & \multicolumn{1}{c|}{5.6}   & 9.9  & \multicolumn{1}{c|}{-3.8} & -5.0 & \multicolumn{1}{c|}{-8.4}  & -13.8 & \multicolumn{1}{c|}{10.1} & 2.4  & \multicolumn{1}{c|}{-4.3} & -10.1 \\
                             & \lmflatness{}                     & \multicolumn{1}{c|}{10.6}  & 6.2  & \multicolumn{1}{c|}{20.1} & 18.1 & \multicolumn{1}{c|}{23.7}  & 23.4  & \multicolumn{1}{c|}{14.2} & 12.0 & \multicolumn{1}{c|}{24.2} & \color[HTML]{FE0000}26.9  \\
                             & MI+Sen                   & \multicolumn{1}{c|}{20.5}  & 19.2 & \multicolumn{1}{c|}{18.2} & 21.0 & \multicolumn{1}{c|}{21.9}  & 22.0  & \multicolumn{1}{c|}{20.1} & 19.7 & \multicolumn{1}{c|}{18.0} & 20.4  \\
                             & MI+\lmflatness{}                  & \multicolumn{1}{c|}{21.4}  & 20.5 & \multicolumn{1}{c|}{22.3} & 19.8 & \multicolumn{1}{c|}{25.1}  & \color[HTML]{FE0000}26.9  & \multicolumn{1}{c|}{22.3} & \color[HTML]{FE0000}20.9 & \multicolumn{1}{c|}{16.4} & 18.1  \\
                             & Sen+\lmflatness{}                 & \multicolumn{1}{c|}{\color[HTML]{FE0000}25.3}  & 21.8 & \multicolumn{1}{c|}{\color[HTML]{FE0000}24.3} & \color[HTML]{FE0000}25.7 & \multicolumn{1}{c|}{\color[HTML]{FE0000}25.3}  & 22.3  & \multicolumn{1}{c|}{\color[HTML]{FE0000}24.1} & 20.0 & \multicolumn{1}{c|}{\color[HTML]{FE0000}25.3} & 25.5  \\ \bottomrule[2pt]
\end{tabular}
\caption{Pearson (Pr) and Spearman (Spr) correlation between prompts' performance and the metrics of various method. Overall, flatness-based metrics obtain higher correlations. {\color[HTML]{FE0000} Red} means the best performance.}
\label{correlation}
\end{table*}

\newpage

\section{Results on Prompt Retrieval}\label{app:ret}
Here are the full results of prompt retrieval performance in our paper, as shown in \autoref{re1} and \autoref{re2}.

\begin{table*}[ht]
\centering
\small
\begin{tabular}{c|c|ccc|ccc|ccc}
\toprule[2pt]
\multirow{2}{*}{Model}       & \multirow{2}{*}{Methods} & \multicolumn{3}{c|}{SST-2}                                         & \multicolumn{3}{c|}{RTE}                                         & \multicolumn{3}{c}{TREC}                                         \\ \cmidrule(l){3-11} 
                             &                          & \multicolumn{1}{c|}{N@1} & \multicolumn{1}{c|}{N@3} & \rate{} & \multicolumn{1}{c|}{N@1} & \multicolumn{1}{c|}{N@3} & \rate{} & \multicolumn{1}{c|}{N@1} & \multicolumn{1}{c|}{N@3} & \rate{} \\ \midrule[2pt]
\multirow{4}{*}{GPT2-base}   & MI                       & \multicolumn{1}{c|}{54.1}   & \multicolumn{1}{c|}{55.0}   & 83.7 & \multicolumn{1}{c|}{56.1}   & \multicolumn{1}{c|}{51.5}   & 87.8 & \multicolumn{1}{c|}{53.6}   & \multicolumn{1}{c|}{54.8}   & 83.9 \\
                             & MI+\lmflatness{}                  & \multicolumn{1}{c|}{\color[HTML]{FE0000}56.6}   & \multicolumn{1}{c|}{59.4}   & 90.8 & \multicolumn{1}{c|}{48.2}   & \multicolumn{1}{c|}{49.4}   & 92.1 & \multicolumn{1}{c|}{55.8}   & \multicolumn{1}{c|}{\color[HTML]{FE0000}56.7}   & 91.2 \\
                             & Sen                      & \multicolumn{1}{c|}{52.2}   & \multicolumn{1}{c|}{53.6}   & 83.1 & \multicolumn{1}{c|}{55.5}   & \multicolumn{1}{c|}{51.3}   & 86.5 & \multicolumn{1}{c|}{53.8}   & \multicolumn{1}{c|}{55.0}   & 83.4 \\
                             & Sen+\lmflatness{}                 & \multicolumn{1}{c|}{49.2}   & \multicolumn{1}{c|}{\color[HTML]{FE0000}59.9}   & \color[HTML]{FE0000}92.7 & \multicolumn{1}{c|}{\color[HTML]{FE0000}69.1}   & \multicolumn{1}{c|}{\color[HTML]{FE0000}57.8}   & \color[HTML]{FE0000}92.4 & \multicolumn{1}{c|}{54.7}   & \multicolumn{1}{c|}{56.0}   & \color[HTML]{FE0000}93.3 \\ \midrule
\multirow{4}{*}{GPT2-medium} & MI                       & \multicolumn{1}{c|}{43.2}   & \multicolumn{1}{c|}{48.7}   & 81.9 & \multicolumn{1}{c|}{43.5}   & \multicolumn{1}{c|}{48.7}   & 85.9 & \multicolumn{1}{c|}{52.9}   & \multicolumn{1}{c|}{57.6}   & 86.0 \\
                             & MI+\lmflatness{}                  & \multicolumn{1}{c|}{51.6}   & \multicolumn{1}{c|}{\color[HTML]{FE0000}52.2}   & \color[HTML]{FE0000}91.8 & \multicolumn{1}{c|}{56.6}   & \multicolumn{1}{c|}{56.2}   & 92.8 & \multicolumn{1}{c|}{\color[HTML]{FE0000}58.0}   & \multicolumn{1}{c|}{\color[HTML]{FE0000}60.1}   & 94.0 \\
                             & Sen                      & \multicolumn{1}{c|}{45.1}   & \multicolumn{1}{c|}{49.0}   & 82.7 & \multicolumn{1}{c|}{44.6}   & \multicolumn{1}{c|}{55.2}   & 88.7 & \multicolumn{1}{c|}{54.0}   & \multicolumn{1}{c|}{56.0}   & 86.4 \\
                             & Sen+\lmflatness{}                 & \multicolumn{1}{c|}{\color[HTML]{FE0000}54.1}   & \multicolumn{1}{c|}{57.9}   & 90.3 & \multicolumn{1}{c|}{57.1}   & \multicolumn{1}{c|}{57.9}   & 93.3 & \multicolumn{1}{c|}{57.4}   & \multicolumn{1}{c|}{57.9}   &\color[HTML]{FE0000} 95.2 \\ \midrule
\multirow{4}{*}{GPT2-large}  & MI                       & \multicolumn{1}{c|}{47.5}   & \multicolumn{1}{c|}{44.9}   & 81.0 & \multicolumn{1}{c|}{36.6}   & \multicolumn{1}{c|}{40.6}   & 90.2 & \multicolumn{1}{c|}{56.1}   & \multicolumn{1}{c|}{54.6}   & 88.7 \\
                             
                             & MI+\lmflatness{}                  & \multicolumn{1}{c|}{51.2}   & \multicolumn{1}{c|}{56.8}   & 90.4 & \multicolumn{1}{c|}{\color[HTML]{FE0000}64.1}   & \multicolumn{1}{c|}{42.3}   & 96.0 & \multicolumn{1}{c|}{\color[HTML]{FE0000}58.1}   & \multicolumn{1}{c|}{54.2}   & 93.4 \\
                             & Sen                      & \multicolumn{1}{c|}{50.2}   & \multicolumn{1}{c|}{51.7}   & 77.5 & \multicolumn{1}{c|}{51.1}   & \multicolumn{1}{c|}{47.5}   & 86.5 & \multicolumn{1}{c|}{49.9}   & \multicolumn{1}{c|}{53.2}   & 87.9 \\
                             & Sen+\lmflatness{}                 & \multicolumn{1}{c|}{\color[HTML]{FE0000}56.4}   & \multicolumn{1}{c|}{\color[HTML]{FE0000}59.0}   &\color[HTML]{FE0000} 91.7 & \multicolumn{1}{c|}{60.9}   & \multicolumn{1}{c|}{\color[HTML]{FE0000}56.3}   &\color[HTML]{FE0000} 97.6 & \multicolumn{1}{c|}{57.0}   & \multicolumn{1}{c|}{\color[HTML]{FE0000}56.8}   &\color[HTML]{FE0000} 94.9 \\ \midrule
\multirow{4}{*}{GPT2-xl}     & MI                       & \multicolumn{1}{c|}{52.1}   & \multicolumn{1}{c|}{51.2}   & 86.1 & \multicolumn{1}{c|}{22.7}   & \multicolumn{1}{c|}{28.4}   & 87.7 & \multicolumn{1}{c|}{54.6}   & \multicolumn{1}{c|}{55.0}   & 85.6 \\
                             
                             & MI+\lmflatness{}                  & \multicolumn{1}{c|}{52.0}   & \multicolumn{1}{c|}{53.5}   & \color[HTML]{FE0000}95.5 & \multicolumn{1}{c|}{\color[HTML]{FE0000}44.6}   & \multicolumn{1}{c|}{\color[HTML]{FE0000}42.6}   &\color[HTML]{FE0000} 97.9 & \multicolumn{1}{c|}{51.4}   & \multicolumn{1}{c|}{53.0}   &\color[HTML]{FE0000} 95.3 \\
                             & Sen                      & \multicolumn{1}{c|}{47.7}   & \multicolumn{1}{c|}{53.1}   & 87.5 & \multicolumn{1}{c|}{23.6}   & \multicolumn{1}{c|}{26.8}   & 86.4 & \multicolumn{1}{c|}{52.8}   & \multicolumn{1}{c|}{53.4}   & 85.0 \\
                             & Sen+\lmflatness{}                 & \multicolumn{1}{c|}{\color[HTML]{FE0000}57.3}   & \multicolumn{1}{c|}{54.2}   & 95.0 & \multicolumn{1}{c|}{32.9}   & \multicolumn{1}{c|}{36.5}   & 96.2 & \multicolumn{1}{c|}{56.6}   & \multicolumn{1}{c|}{53.4}   & 95.0 \\ \bottomrule[2pt]
\end{tabular}
\caption{Results of high-performance prompts retrieval, we can see that metric combined prompt loss and flatness achieve better performance. Specifically, N@1 represents NDCG@1. {\color[HTML]{FE0000} Red} means the best performance.}
\label{re1}
\end{table*}

\begin{table*}[!ht]\centering
\small
\begin{tabular}{c|c|ccc|ccc|ccc}
\toprule[2pt]
\multirow{2}{*}{Model}       & \multirow{2}{*}{Methods} & \multicolumn{3}{c|}{AGNews}                                      & \multicolumn{3}{c|}{CB}                                          & \multicolumn{3}{c}{DBpedia}                                      \\ \cmidrule(l){3-11} 
                             &                          & \multicolumn{1}{c|}{N@1} & \multicolumn{1}{c|}{N@3} & \rate{} & \multicolumn{1}{c|}{N@1} & \multicolumn{1}{c|}{N@3} & \rate{} & \multicolumn{1}{c|}{N@1} & \multicolumn{1}{c|}{N@3} & \rate{} \\ \midrule[2pt]
\multirow{4}{*}{GPT2-base}   & MI                       & \multicolumn{1}{c|}{49.2}   & \multicolumn{1}{c|}{50.4}   & 86.8 & \multicolumn{1}{c|}{31.5}   & \multicolumn{1}{c|}{42.6}   & 88.2 & \multicolumn{1}{c|}{39.9}   & \multicolumn{1}{c|}{47.8}   & 86.6 \\
                             & Sen                      & \multicolumn{1}{c|}{46.5}   & \multicolumn{1}{c|}{56.8}   & 85.4 & \multicolumn{1}{c|}{34.3}   & \multicolumn{1}{c|}{50.0}   & 85.6 & \multicolumn{1}{c|}{35.9}   & \multicolumn{1}{c|}{46.3}   & 86.7 \\
                             & MI+\lmflatness{}                  & \multicolumn{1}{c|}{52.1}   & \multicolumn{1}{c|}{48.9}   & 91.9 & \multicolumn{1}{c|}{\color[HTML]{FE0000}43.1}   & \multicolumn{1}{c|}{44.3}   & 94.6 & \multicolumn{1}{c|}{39.9}   & \multicolumn{1}{c|}{\color[HTML]{FE0000}52.8}   & 94.2 \\
                             & Sen+\lmflatness{}                 & \multicolumn{1}{c|}{\color[HTML]{FE0000}52.3}   & \multicolumn{1}{c|}{54.0}   & \color[HTML]{FE0000}92.8 & \multicolumn{1}{c|}{34.8}   & \multicolumn{1}{c|}{45.4}   & \color[HTML]{FE0000}95.1 & \multicolumn{1}{c|}{50.1}   & \multicolumn{1}{c|}{48.4}   & \color[HTML]{FE0000}94.8 \\ \midrule
\multirow{4}{*}{GPT2-medium} & MI                       & \multicolumn{1}{c|}{46.8}   & \multicolumn{1}{c|}{50.4}   & 88.2 & \multicolumn{1}{c|}{57.8}   & \multicolumn{1}{c|}{48.5}   & 86.2 & \multicolumn{1}{c|}{51.3}   & \multicolumn{1}{c|}{50.0}   & 86.6 \\
                             & Sen                      & \multicolumn{1}{c|}{44.3}   & \multicolumn{1}{c|}{56.8}   & 85.9 & \multicolumn{1}{c|}{64.4}   & \multicolumn{1}{c|}{60.6}   & 86.0 & \multicolumn{1}{c|}{53.3}   & \multicolumn{1}{c|}{52.0}   & 87.1 \\
                             & MI+\lmflatness{}                  & \multicolumn{1}{c|}{\color[HTML]{FE0000}51.9}   & \multicolumn{1}{c|}{\color[HTML]{FE0000}58.9}   & 93.9 & \multicolumn{1}{c|}{53.7}   & \multicolumn{1}{c|}{47.0}   & \color[HTML]{FE0000}95.7 & \multicolumn{1}{c|}{51.3}   & \multicolumn{1}{c|}{\color[HTML]{FE0000}58.4}   & 95.2 \\
                             & Sen+\lmflatness{}                 & \multicolumn{1}{c|}{50.8}   & \multicolumn{1}{c|}{54.0}   & \color[HTML]{FE0000}94.8 & \multicolumn{1}{c|}{\color[HTML]{FE0000}63.7}   & \multicolumn{1}{c|}{52.0}   & 94.5 & \multicolumn{1}{c|}{\color[HTML]{FE0000}56.0}   & \multicolumn{1}{c|}{55.0}   & \color[HTML]{FE0000}95.6 \\ \midrule
\multirow{4}{*}{GPT2-large}  & MI                       & \multicolumn{1}{c|}{\color[HTML]{FE0000}53.2}   & \multicolumn{1}{c|}{51.2}   & 87.3 & \multicolumn{1}{c|}{34.1}   & \multicolumn{1}{c|}{44.8}   & 85.9 & \multicolumn{1}{c|}{53.3}   & \multicolumn{1}{c|}{55.3}   & 87.0 \\
                             & Sen                      & \multicolumn{1}{c|}{46.9}   & \multicolumn{1}{c|}{50.6}   & 84.0 & \multicolumn{1}{c|}{28.8}   & \multicolumn{1}{c|}{40.1}   & 83.1 & \multicolumn{1}{c|}{33.8}   & \multicolumn{1}{c|}{43.1}   & 84.8 \\
                             & MI+\lmflatness{}                  & \multicolumn{1}{c|}{47.0}   & \multicolumn{1}{c|}{45.9}   & \color[HTML]{FE0000}95.2 & \multicolumn{1}{c|}{\color[HTML]{FE0000}37.1}   & \multicolumn{1}{c|}{\color[HTML]{FE0000}50.1}   &\color[HTML]{FE0000} 96.3 & \multicolumn{1}{c|}{\color[HTML]{FE0000}50.9}   & \multicolumn{1}{c|}{49.1}   & 96.3 \\
                             & Sen+\lmflatness{}                 & \multicolumn{1}{c|}{52.1}   & \multicolumn{1}{c|}{53.8}   & 94.7 & \multicolumn{1}{c|}{24.1}   & \multicolumn{1}{c|}{46.2}   & 95.9 & \multicolumn{1}{c|}{39.6}   & \multicolumn{1}{c|}{54.8}   & \color[HTML]{FE0000}97.1 \\ \midrule
\multirow{4}{*}{GPT2-xl}     & MI                       & \multicolumn{1}{c|}{44.5}   & \multicolumn{1}{c|}{60.8}   & 88.6 & \multicolumn{1}{c|}{51.4}   & \multicolumn{1}{c|}{62.3}   & 86.1 & \multicolumn{1}{c|}{55.7}   & \multicolumn{1}{c|}{53.0}   & 85.7 \\
                             & Sen                      & \multicolumn{1}{c|}{48.1}   & \multicolumn{1}{c|}{57.0}   & 87.8 & \multicolumn{1}{c|}{48.8}   & \multicolumn{1}{c|}{53.0}   & 83.1 & \multicolumn{1}{c|}{44.1}   & \multicolumn{1}{c|}{52.9}   & 84.1 \\
                             & MI+\lmflatness{}                  & \multicolumn{1}{c|}{48.9}   & \multicolumn{1}{c|}{46.3}   & \color[HTML]{FE0000}97.4 & \multicolumn{1}{c|}{\color[HTML]{FE0000}53.6}   & \multicolumn{1}{c|}{\color[HTML]{FE0000}69.2}   &\color[HTML]{FE0000} 96.4 & \multicolumn{1}{c|}{\color[HTML]{FE0000}58.7}   & \multicolumn{1}{c|}{49.7}   & 96.0 \\
                             & Sen+\lmflatness{}                 & \multicolumn{1}{c|}{53.0}   & \multicolumn{1}{c|}{57.1}   & 96.0 & \multicolumn{1}{c|}{54.8}   & \multicolumn{1}{c|}{54.1}   & 96.0 & \multicolumn{1}{c|}{47.7}   & \multicolumn{1}{c|}{58.4}   & \color[HTML]{FE0000}96.2 \\ \bottomrule[2pt]
\end{tabular}
\caption{Results of high-performance prompts retrieval on AGNews, CB, and DBpedia. we can see that metric combined prompt loss and flatness achieve better performance.  {\color[HTML]{FE0000} Red} means the best performance.}
\label{re2}
\end{table*}

\newpage
\section{Comparison to Automatic Prompt Generation Algorithms}\label{RLPrompt}
Here we compare \lmflatness{} to automatic prompt-generation, namely RLPrompt \cite{deng2022rlprompt}, Tempera \cite{zhang2023tempera}, and GrIPs \cite{prasad2022grips}. 
The primary objective of these algorithms is to automatically generate prompts that would be apt for prompt tuning. By contrast, our study aims to scrutinize and identify a prompt that would be advantageous for ICL. In this section, we present empirical results based on prompts produced by off-the-shelf models of RLPrompt, Tempera, GrIPs. Then we compare the performance of the prompts obtained via various approaches, including RLPrompt, Tempera, GrIPs, and our method (Sen+\lmflatness{}), as depicted in \autoref{rlprompt}. The results illustrate that the prompt selected by our method exhibits superior ICL performance. Besides, we show some examples of RLPrompt, Tempera, and GrIPs in \autoref{prompts:example:sst},\autoref{prompts:example:rte},\autoref{prompts:example:trec}.

\begin{table*}[!ht]\centering\small
\begin{tabular}{c|c|ccc|ccc|ccc}
\toprule[2pt]
\multirow{2}{*}{Model}       & \multirow{2}{*}{Methods} & \multicolumn{3}{c|}{SST-2}                                         & \multicolumn{3}{c|}{RTE}                                         & \multicolumn{3}{c}{TREC}                                         \\ \cmidrule(l){3-11} 
                             &                          & \multicolumn{1}{c|}{1-shot} & \multicolumn{1}{c|}{4-shot} & 8-shot & \multicolumn{1}{c|}{1-shot} & \multicolumn{1}{c|}{4-shot} & 8-shot & \multicolumn{1}{c|}{1-shot} & \multicolumn{1}{c|}{4-shot} & 8-shot \\ \midrule[2pt]
\multirow{4}{*}{GPT2-xl}   & RLPrompt                       & \multicolumn{1}{c|}{54.1}   & \multicolumn{1}{c|}{56.0}   & 60.7 & \multicolumn{1}{c|}{52.1}   & \multicolumn{1}{c|}{54.5}   & 57.8 & \multicolumn{1}{c|}{25.6}   & \multicolumn{1}{c|}{27.8}   & 30.9 \\
                             & Tempera                  & \multicolumn{1}{c|}{55.0}   & \multicolumn{1}{c|}{59.4}   & 61.8 & \multicolumn{1}{c|}{52.2}   & \multicolumn{1}{c|}{58.4}   & 59.1 & \multicolumn{1}{c|}{24.8}   & \multicolumn{1}{c|}{28.7}   & 31.2 \\
                             & GrIPs                      & \multicolumn{1}{c|}{52.2}   & \multicolumn{1}{c|}{58.6}   & 60.1 & \multicolumn{1}{c|}{51.5}   & \multicolumn{1}{c|}{53.3}   & 56.5 & \multicolumn{1}{c|}{23.8}   & \multicolumn{1}{c|}{25.0}   & 27.9 \\
                             & Sen+\lmflatness{}                 & \multicolumn{1}{c|}{\color[HTML]{FE0000}58.9}   & \multicolumn{1}{c|}{\color[HTML]{FE0000}63.9}   & \color[HTML]{FE0000}65.7 & \multicolumn{1}{c|}{\color[HTML]{FE0000}55.6}   & \multicolumn{1}{c|}{\color[HTML]{FE0000}58.9}   & \color[HTML]{FE0000}61.9 & \multicolumn{1}{c|}{\color[HTML]{FE0000}29.7}   & \multicolumn{1}{c|}{\color[HTML]{FE0000}32.1}   & \color[HTML]{FE0000}34.7 \\  \bottomrule[2pt]
\end{tabular}
\caption{In-context learning performance of prompts from different methods. We can observe that the prompt selected by our method achieves better in-context learning performance.}
\label{rlprompt}
\end{table*}

\begin{table*}[ht]
    \centering
    \small
    \noindent\fbox{%
    \begin{minipage}{\dimexpr\linewidth-2\fboxsep-2\fboxrule} 
\paragraph{Examples of prompt generated by RLPrompt, Tempera, and GrIPs}
\begin{itemize}
    \item \textbf{[RLPrompt]}: Sentiment of the sentence is negative or positive.

    \item \textbf{[Tempera]}: Given text, given text, Classify whether it is good or bad.
    \item \textbf{[GrIPs]}: Your task as "positive" or "negative".
\end{itemize}
    \end{minipage}
}
    \caption{Instructions from RLPrompt, Tempera, and GrIPs for SST-2 task}
    \label{prompts:example:sst}
\end{table*}

\begin{table*}[ht]
    \centering
    \small
    \noindent\fbox{%
    \begin{minipage}{\dimexpr\linewidth-2\fboxsep-2\fboxrule} 
\paragraph{Examples of prompt generated by RLPrompt, Tempera, and GrIPs}
\begin{itemize}
    \item \textbf{[RLPrompt]}: premise follow that hypo yes or no?

    \item \textbf{[Tempera]}: Given premise, does it follow hypothesis?
    \item \textbf{[GrIPs]}: Does the information support premise?
\end{itemize}
    \end{minipage}
}
    \caption{Instructions from RLPrompt, Tempera, and GrIPs for RTE task}
    \label{prompts:example:rte}
\end{table*}

\begin{table*}[!h]
    \centering
    \small
    \noindent\fbox{%
    \begin{minipage}{\dimexpr\linewidth-2\fboxsep-2\fboxrule} 
\paragraph{Examples of prompt generated by RLPrompt, Tempera, and GrIPs}
\begin{itemize}
    \item \textbf{[RLPrompt]}: The topic of the question is

    \item \textbf{[Tempera]}: Given the info, what's the topic
    \item \textbf{[GrIPs]}: Topic of the sentence
\end{itemize}
    \end{minipage}
}
    \caption{Instructions from RLPrompt, Tempera, and GrIPs for TREC task}
    \label{prompts:example:trec}
\end{table*}

\newpage

\section{Instructions}
\label{app:discrete-prompt}
Here we include the pool of natural language prompts (instructions) used in each task. We list instructions for SST-2 in Table~\ref{prompts:sst}, RTE in Table~\ref{prompts:rte}, TREC in Table~\ref{prompts:trec}, AGNews in Table~\ref{prompts:agnews}, CB in Table~\ref{prompts:cb} and DBPedia in Table~\ref{prompts:dbpedia}.

\begin{table*}[ht]
    \centering
    \small
    \noindent\fbox{%
    \begin{minipage}{\dimexpr\linewidth-2\fboxsep-2\fboxrule} 
\paragraph{SST-2 Instructions}
\begin{itemize}
    \item Suppose we have the following premise, Can we infer that hypothesis? Yes, no, or maybe?
    \item Based on the previous premise, is it true for the hypothesis?
    \item See on the following information, is the claim right?
    \item Given that premise, does it follow that hypothesis? Yes, no, or maybe?
    \item Given the premise, are we justified in saying that hypothesis? Yes, no, or maybe?
    \item Based on the text, question: hypothesis is True, False, or Neither?
    \item Keeping in mind the above text, consider: hypothesis  is always, sometimes, or never correct?
    \item Given premise. Is it guaranteed true that hypothesis? Yes, no, or maybe?
    \item Given that premise. Therefore, it must be true that hypothesis? Yes, no, or maybe?
    \item Assume it is true that premise. Therefore, hypothesis is guaranteed, possible, or impossible?
    \item Using only the following description and what you know about the world, hypothesis is definitely correct, incorrect, or inconclusive?
    \item Take the following as truth. Then the hypothesis is true, false, or inconclusive?
    \item Can we derive that hypothesis if we have the following premise? Yes, no, or perhaps?
    \item Can we arrive at that conclusion if we possess the following information? Possibly, no, or both?
    \item Does that premise flow from the given premise? Yes, no, or perhaps?
    \item Does that information support the claim?
    \item Is the assertion accurate in light of such information?
    \item Considering the text, which of the following statements is True, False, or Both?
    \item Think about the question: Is hypothesis always, occasionally, or never correct?
    \item Can we derive that conclusion if we have the following information? Yes, no, or possibly?
\end{itemize}
    \end{minipage}
}
    \caption{Instructions for SST-2 task}
    \label{prompts:sst}
\end{table*}

\begin{table*}[ht]
    \centering
    \small
    \noindent\fbox{%
    \begin{minipage}{\dimexpr\linewidth-2\fboxsep-2\fboxrule} 
\paragraph{RTE Instructions}
\begin{itemize}
    \item Using only the above description and what you know about the world, is hypothesis definitely correct? Yes or no?
    \item Given premise, Is it guaranteed true that hypothesis? Yes or no?
    \item Suppose premise, Can we infer that hypothesis? Yes or no?
    \item Given premise Should we assume that hypothesis is true? Yes or no?
    \item Given that premise, Does it follow that hypothesis Yes or no?
    \item Given premise. Is it guaranteed true that hypothesis? Yes, no, or maybe?
    \item Given that premise. Therefore, it must be true that hypothesis? Yes, no, or maybe?
    \item Assume it is true that premise. Therefore, hypothesis is guaranteed, possible, or impossible?
    \item Using only the following description and what you know about the world, hypothesis is definitely correct, incorrect, or inconclusive?
    \item Take the following as truth. Then the hypothesis is true, false, or inconclusive?
    \item Can we derive that hypothesis if we have the following premise? Yes, no, or perhaps?
    \item Can we arrive at that conclusion if we possess the following information? Possibly, no, or both?
    \item Does that premise flow from the given premise? Yes, no, or perhaps?
    \item Does that information support the claim?
    \item Is the assertion accurate in light of such information?
    \item Considering the text, which of the following statements is True, False, or Both?
    \item Think about the question: Is hypothesis always, occasionally, or never correct?
    \item Can we derive that conclusion if we have the following information? Yes, no, or possibly?
    \item Suppose we have the following premise, Can we infer that hypothesis? Yes, no, or maybe?
    \item Based on the previous premise, is it true for the hypothesis?
\end{itemize}
    \end{minipage}
}
    \caption{Instructions for RTE task}
    \label{prompts:rte}
\end{table*}

\begin{table*}[ht]
    \centering
    \small
    \noindent\fbox{%
    \begin{minipage}{\dimexpr\linewidth-2\fboxsep-2\fboxrule} 
\paragraph{TREC Instructions}
\begin{itemize}
    \item What kind of label best describes this question below?
    \item What is this a piece of question regarding for?
    \item What is the category of the following question?
    \item Which is the most relevant topic of the following question?
    \item Give the topic of the given question.
    \item Read the question below, provide its focused topic.
    \item Is this a piece of question regarding ABBR, ENTY, DESC, HUM, LOC, or NUM?
    \item Which section of a newspaper would this question likely appear in?
    \item What label would you use to characterize this question item?
    \item What term can best sums up this question?
    \item Which category most accurately sums up this question item?
    \item What label would you use to characterize this question?
    \item Is this question related to ABBR, ENTY, DESC, HUM, LOC, or NUM?
    \item Does this question story have anything to do with ABBR, ENTY, DESC, HUM, LOC, or NUM?
    \item Read the question below and explain its specific subject.
    \item Please read the following material and explain its main point.
    \item Provide your thoughts on the content below after reading it.
    \item Describe the question's subject as follows.
    \item For what purpose does this question item exist?
    \item Are there any ABBR, ENTY, DESC, HUM, LOC, or NUM related stories in this question?
\end{itemize}
    \end{minipage}
}
    \caption{Instructions for TREC task}
    \label{prompts:trec}
\end{table*}

\begin{table*}[ht]
    \centering
    \small
    \noindent\fbox{%
    \begin{minipage}{\dimexpr\linewidth-2\fboxsep-2\fboxrule} 
\paragraph{AGNews Instructions}
\begin{itemize}
    \item What label best describes this news article?
    \item What is this a piece of news regarding for?
    \item What is the category of the following news?
    \item Which is the most relevant topic of the following news?
    \item Give the topic of the given text.
    \item Read the text below, provide its focused topic.
    \item Is this a piece of news regarding world, sport, business,or science?
    \item Which section of a newspaper would this article likely appear in?
    \item What label would you use to characterize this news item?
    \item What term best sums up this news report?
    \item Which category most accurately sums up this news item?
    \item What label would you use to characterize this news story?
    \item Is this news related to the world, sports, business, or science?
    \item Does this news story have anything to do with the world, sports, business, or science?
    \item Read the paragraph below and explain its specific subject.
    \item Please read the following material and explain its main point.
    \item Provide your thoughts on the content below after reading it.
    \item Describe the text's subject as follows.
    \item For what purpose does this news item exist?
    \item Are there any world-related, sports, business, or science-related stories in this news?
\end{itemize}
    \end{minipage}
}
    \caption{Instructions for AGNews task}
    \label{prompts:agnews}
\end{table*}

\begin{table*}[ht]
    \centering
    \small
    \noindent\fbox{%
    \begin{minipage}{\dimexpr\linewidth-2\fboxsep-2\fboxrule} 
\paragraph{CB Instructions}
\begin{itemize}
    \item Suppose we have the following premise, Can we infer that hypothesis? Yes, no, or maybe?
    \item Based on the previous premise, is it true for the hypothesis?
    \item See on the following information, is the claim right?
    \item Given that premise, does it follow that hypothesis? Yes, no, or maybe?
    \item Given the premise, are we justified in saying that hypothesis? Yes, no, or maybe?
    \item Based on the text, question: hypothesis is True, False, or Neither?
    \item Keeping in mind the above text, consider: hypothesis  is always, sometimes, or never correct?
    \item Given premise. Is it guaranteed true that hypothesis? Yes, no, or maybe?
    \item Given that premise. Therefore, it must be true that hypothesis? Yes, no, or maybe?
    \item Assume it is true that premise. Therefore, hypothesis is guaranteed, possible, or impossible?
    \item Using only the following description and what you know about the world, hypothesis is definitely correct, incorrect, or inconclusive?
    \item Take the following as truth. Then the hypothesis is true, false, or inconclusive?
    \item Can we derive that hypothesis if we have the following premise? Yes, no, or perhaps?
    \item Can we arrive at that conclusion if we possess the following information? Possibly, no, or both?
    \item Does that premise flow from the given premise? Yes, no, or perhaps?
    \item Does that information support the claim?
    \item Is the assertion accurate in light of such information?
    \item Considering the text, which of the following statements is True, False, or Both?
    \item Think about the question: Is hypothesis always, occasionally, or never correct?
    \item Can we derive that conclusion if we have the following information? Yes, no, or possibly?
\end{itemize}
    \end{minipage}
}
    \caption{Instructions for CB task}
    \label{prompts:cb}
\end{table*}

\begin{table*}[ht]
    \centering
    \small
    \noindent\fbox{%
    \begin{minipage}{\dimexpr\linewidth-2\fboxsep-2\fboxrule} 
\paragraph{DBPedia Instructions}
\begin{itemize}
    \item What label best describes this paragraph?
    \item What is this paragraph regarding for?
    \item What is the category of the following paragraph?
    \item Which is the most relevant topic of the following paragraph?
    \item Give the topic of the given text.
    \item Read the text below, provide its focused topic.
    \item Is this paragraph regarding company, educational institution, artist, athlete, office holder, mean of transportation, building, natural place, village, animal, plant, album, film or written work?
    \item What label would you use to characterize this paragraph?
    \item What term best sums up this paragraph?
    \item Which category most accurately sums up this paragraph?
    \item What label would you use to characterize this paragraph?
    \item Is this paragraph related to company, educational institution, artist, athlete, office holder, mean of transportation, building, natural place, village, animal, plant, album, film or written work?
    \item Does this news story have anything to do with company, educational institution, artist, athlete, office holder, mean of transportation, building, natural place, village, animal, plant, album, film or written work?
    \item Read the paragraph below and explain its specific subject.
    \item Please read the following material and explain its main point.
    \item Describe the text's subject as follows.
    \item Are there any company, educational institution, artist, athlete, office holder, mean of transportation, building, natural place, village, animal, plant, album, film or written work content in this paragraph?
    \item Given a list of categories: company, educational institution, artist, athlete, office holder, mean of transportation, building, natural place, village, animal, plant, album, film or written work, what category does the paragraph belong to?
    \item Pick one category for the following text. The options are - company, educational institution, artist, athlete, office holder, mean of transportation, building, natural place, village, animal, plant, album, film or written work.
    \item Given a choice of categories company, educational institution, artist, athlete, office holder, mean of transportation, building, natural place, village, animal, plant, album, film or written work, the text refers to which one?
\end{itemize}
    \end{minipage}
}
    \caption{Instructions for DBPedia task}
    \label{prompts:dbpedia}
\end{table*}

\end{document}